\documentclass[journal]{IEEEtran}

\usepackage[pdftex]{graphicx}
\usepackage{amssymb}
\usepackage{booktabs}
\usepackage{array}
\usepackage[cmex10]{amsmath}

\usepackage{url}
\usepackage{cite}

\usepackage{stfloats}
\usepackage{epstopdf}
\usepackage{cite}
\usepackage{verbatim}

\usepackage[protrusion=true,expansion=true]{microtype}

\DeclareMathOperator{\TV}{TV}
\DeclareMathOperator{\TVD}{tvd}

\renewcommand{\le}{ \leqslant }
\renewcommand{\ge}{ \geqslant }

\providecommand{\norm}[1]{\lVert#1\rVert}
\providecommand{\twonorm}[1]{\left \Vert #1 \right \Vert_{2}}
\providecommand{\smalltwonorm}[1]{\| #1 \|_{2}}

\providecommand{\onenorm}[1]{\lVert#1\rVert_{1}}

\newcommand{\sumK}{ \sum_{k = 0}^{K-1} }
\newcommand{\sumN}{ \sum_{n = 0}^{N-1} }

\newcommand{\ST}{\text{s. t. \ }}

\newcommand{\DD}{\mathbf{D}}

\newcommand{\wM}{\omega_{M}}

\newcommand{\dw}{\Delta \omega}

\newcommand{\x}{\mathbf{x}}
\newcommand{\y}{\mathbf{y}}

\renewcommand{\a}{\mathbf{a}}
\renewcommand{\b}{\mathbf{b}}

\newcommand{\ak}{\mathbf{a}_k}
\newcommand{\bk}{\mathbf{b}_k}

\newcommand{\ank}{a(  n  ,  k  )}
\newcommand{\bnk}{b(  n  ,  k  )}

\newcommand{\cnk}{c(  n  ,  k  )}
\newcommand{\snk}{s(  n  ,  k  )}

\renewcommand{\u}{\mathbf{u}}
\renewcommand{\v}{\mathbf{v}}

\newcommand{\w}{\mathbf{w}}

\newcommand{\unk}{u(n,k)}

\newcommand{\uk}{\mathbf{u}_{k}}

\newcommand{\uink}{u^{(i)}(  n  ,  k  )}
\newcommand{\vink}{v^{(i)}(  n  ,  k  )}

\newcommand{\vnk}{v(n,k)}

\newcommand{\vk}{\mathbf{v}_{k}}

\newcommand{\p}{\mathbf{p}}
\newcommand{\q}{\mathbf{q}}

\newcommand{\pnk}{p(n,k)}
\newcommand{\qnk}{q(n,k)}

\newcommand{\pk}{\mathbf{p}_k}
\newcommand{\qk}{\mathbf{q}_k}

\allowdisplaybreaks

\newcommand{\figurescale}{0.45}

\newcommand{\ia}{({\it i\/})}
\newcommand{\ib}{({\it ii\/})}

\newcommand{\ZZ}{\mathbb{Z}}
\newcommand{\RR}{\mathbb{R}}

\newcommand{\z}{\mathbf{z}}

\newcommand{\myJ}{ \mathrm{j} }
\newcommand{\myE}{ \mathrm{e} }
\newcommand{\om}{\omega}
\newcommand{\lam}{{\lambda} }

\newcommand{\half}{\frac{1}{2}}
\newcommand{\inv}{^{-1}}

\providecommand{\abs}[1]{\lvert#1\rvert}

\newcommand{\iter}[1]{^{(#1)}}

\newcommand{\conj}[1]{{#1}^{\ast}}			

\usepackage{setspace}

\title{Sparse Frequency Analysis with Sparse-Derivative Amplitude and Phase Functions}
\title{Sparse Frequency Analysis with Sparse-Derivative Instantaneous Amplitude and Phase Functions}
\author{Yin Ding and Ivan~W.~Selesnick%
\thanks{The authors are with the Department of Electrical and Computer Engineering, Polytechnic Institute of New York University, 6 MetroTech Center,  Brooklyn, NY 11201.
Email: adamding0215@me.com and selesi@poly.edu.
Phone: 718 260-3416. Fax: 718 260-3906. }
\thanks{This research was support by the NSF under Grant No. CCF-1018020.}
		}

\begin{document}
\maketitle

\begin{abstract}

This paper addresses the problem of expressing a signal as a sum of frequency components (sinusoids)
wherein each sinusoid may exhibit abrupt changes in its amplitude and/or phase.
The Fourier transform of a narrow-band signal, with a discontinuous amplitude and/or phase function,
exhibits spectral and temporal spreading.
The proposed method aims to avoid such spreading by explicitly modeling
the signal of interest as a sum of sinusoids with time-varying amplitudes.
So as to accommodate abrupt changes, it is further assumed that the 
amplitude/phase functions are approximately piecewise constant (i.e., their time-derivatives are sparse).
The proposed method is based on a convex variational (optimization) approach wherein the
total variation (TV) of the amplitude functions are regularized subject
to a perfect (or approximate) reconstruction constraint.
A computationally efficient algorithm is derived based on convex optimization techniques.
The proposed technique can be used to perform band-pass filtering that is relatively insensitive to 
narrow-band amplitude/phase jumps present in data, which normally 
pose a challenge (due to transients, leakage, etc.).
The method is illustrated using both synthetic signals and human EEG data
for the purpose of band-pass filtering and the estimation of phase synchrony indexes.

\end{abstract}


\section{Introduction}

\IEEEPARstart{S}{everal}
methods in time series analysis aim to quantify the phase behavior of one or more signals 
(e.g., studies of phase synchrony and coherence among EEG channels 
\cite{Lachaux_1999,Aviyente_2011,Almeida_2011,Yan_1999,Hung_Yi_2011,Sanqing_2010}).
These methods are most meaningfully applied to signals that consist primarily of a single narrow-band component.
However, in practice, available data often does not have a frequency spectrum localized to a narrow band,
in which case the data is usually band-pass filtered to isolate the frequency band of interest (e.g., \cite{Hung_Yi_2011,Lachaux_1999}). 
Yet, the process of band-pass filtering has the effect of spreading abrupt changes 
in phase and amplitude across both time and frequency. 
Abrupt changes present in the underlying 
component will be reduced.
This limitation of linear time-invariant (LTI) filtering is well recognized.
In linear signal analysis, time and frequency resolution can be traded-off with one another (c.f. wavelet transforms); 
however, they can not both be improved arbitrarily. Circumventing this limitation requires some form of nonlinear signal analysis
\cite{Chen_1998_SIAM}.
The nonlinear analysis method developed in this paper
is motivated by the problem of extracting (isolating) a narrow-band signal from a generic signal while preserving abrupt 
phase-shifts and amplitude step-changes due for example to phase-resetting \cite{Tass_2007}.

This paper addresses the problem of expressing a signal as a sum of frequency components (sinusoids)
wherein each sinusoid may exhibit abrupt changes in its amplitude and/or phase.
The discrete Fourier transform (DFT) and Fourier series each give a representation of a generic signal as a sum of sinusoids,
wherein the amplitude and phase of each sinusoid is a constant value (not time-varying).
In contrast, in the proposed method,
the amplitude and phase of each sinusoid is a time-varying function.
So as to effectively represent abrupt changes,
it is assumed that the amplitude and phase of each sinusoid is approximately constant;
i.e., the time-derivative of the amplitude and phase of each sinusoid is sparse.
It is further assumed that the signal under analysis admits a sparse frequency representation;
i.e., the signal consists of relatively few frequency components (albeit with time-varying amplitudes and phases).

In the proposed sparse frequency analysis (SFA) approach, the amplitude and phase of each sinusoid is allowed to vary
so as 
\ia\
to match the behavior of data wherein it is known or expected that abrupt changes in
amplitude and phase may be present (e.g., signals with phase-reset phenomena), and
\ib\
to obtain a more sparse signal representation relative to the DFT or oversampled DFT,
thereby improving frequency and phase resolution.

The proposed method has a parameter by which one can tune the behavior of the decomposition.
At one extreme, the method coincides with the DFT.
Hence, the method can be interpreted as a generalization of the DFT.

In order to achieve the above-described signal decomposition, we formulate 
it as the solution to a convex sparsity-regularized linear inverse problem.
Specifically, the total variation (TV) of the amplitude of each sinusoid is regularized.
Two forms of the problem are described.
In the first form, perfect reconstruction is enforced;
in the second form,  the energy of the residual is minimized.
The second form is suitable when the given data is contaminated by additive noise. 
The two forms are analogous to basis pursuit (BP) and basis pursuit denoising (BPD) respectively \cite{Chen_1998_SIAM}.

To solve the formulated optimization problems, 
iterative algorithms are derived using convex optimization techniques:
variable splitting,
the alternating direction method of multipliers (ADMM) \cite{Afonso_TIP_CSALSA, Combettes_2010_chap_B, Boyd_2011_admm},
and majorization-minimization (MM) \cite{FBDN_2007_TIP}.
The developed algorithms use total variation denoising (TVD) \cite{ROF_1992, Chan_2001_TIP} as a sub-step,
which is solved exactly using the recently developed algorithm by Condat \cite{Condat_2012}. 
The resulting algorithm is `matrix-free' in that it does not require solving systems of linear equations
nor accessing individual rows/columns of matrices.

The proposed signal decomposition can be used to perform mildly nonlinear band-pass filtering that is better able to track abrupt amplitude/phase jumps than an LTI bandpass filter.
Such band-pass filtering is also relatively insensitive to narrow-band amplitude/phase jumps present outside the frequency band of interest (interference), which normally pose a challenge (due to transients, leakage, etc.).
The method is illustrated using both synthetic signals and human EEG data, for the purpose of band-pass filtering and the estimation of phase synchrony indexes.
In the examples, the proposed method is compared with the DFT and with band-pass filtering. 
The examples demonstrate the improved ability to represent and detect abrupt phase shifts and amplitude changes when they are present in the data.

\subsection{Related work}

The short-time Fourier transform (STFT) can already be used to some extent to estimate and track the instantaneous 
frequency and amplitude of narrow-band components of a generic signal \cite{Allen_1977_PIEEE, Loughlin_1994, Fulop_2006_JASA};
however, the STFT is computed as a linear transform and is defined in terms of pre-defined basis functions,
whereas the proposed representation is computed as the solution to  a non-quadratic optimization problem. 
As discussed in\cite{Pitton_1996}, the time-frequency resolution of the STFT is constrained by the utilized window. 
Hence, although the STFT can be computed with far less computation, it does not have the properties of the proposed approach. 
On the other hand, the effectiveness and suitability of the proposed approach depends on the validity of the sparsity assumption. 

We also note that the sparse STFT (i.e. basis pursuit with the STFT) and other sparse time-frequency distributions can 
overcome some of the time-frequency limitations of the linear STFT \cite{Flandrin_Goncalves_2004}. 
However, the aim therein is an accurate high resolution time-frequency distribution, whereas the aim of this paper is to 
produce a representation in the time-domain as a sum of sinusoids with abrupt amplitude/phase shifts. 

Sinusoidal modeling also seeks to represent a signal as a sum of sinusoids, each with time-varying amplitude and phase/frequency
\cite{McAulay_1986, Macon_1997_TSAP, Pantazis_2008_ICASSP, Pantazis_2010_SPL}.
However, sinusoidal models usually assume that these are slowly varying functions. 
Moreover, these functions are often parameterized and the parameters estimated through nonlinear least squares or by 
short-time Fourier analysis. 
The goal of this paper is somewhat similar to sinusoidal modeling, yet the model and optimization approach are quite distinct. 
The method presented here is non-parametric and can be understood as a generalization of the DFT.

Empirical mode decomposition (EMD) \cite{Huang_1998_EMD}, 
a nonlinear data-adaptive approach that decomposes a non-stationary signal into oscillatory components, 
also seeks to overcome limitations of linear short-time Fourier analysis 
\cite{Adam_2006_JASA, Rilling_2003_cnf, Rilling_2008_TSP}. 
In EMD, the oscillatory components, or `intrinsic mode functions' (IMFs),
are not restricted to occupy distinct bands.
Hence, EMD provides a form of time-frequency analysis.
In contrast, the sparse frequency analysis (SFA) approach presented
here is more restrictive in that it assumes the narrow band components
occupy non-overlapping frequency bands.
EMD, however, is susceptible to a `mode mixing' issue, wherein the instantaneous 
frequency of an IMF does not accurately track the frequency of the underlying non-stationary component.
This is the case, in particular, when the instantaneous amplitude/phase of a component exhibits an abrupt shift.
Several optimization-based variations and extensions of EMD have been formulated
that aim to provide a more robust and flexible form of EMD
\cite{Meignen_2007_TSP, Peng_2010_TSP, Oberlin_2012_TSP, Pustelnik_2012_EUSIPCO,Hou_2011_AADA}.
In these methods, as in EMD, one narrow-band component (IMF) is extracted at a time,
the frequency support of each component is not constrained a priori,
and the instantaneous amplitude/phase of each component is modeled as slowly varying.
In contrast, the SFA method developed below, optimizes all components jointly, uses
a fixed uniform discretization of frequency, 
and models the instantaneous amplitude/phase as possessing discontinuities.
Hence, SFA and EMD-like methods have somewhat different aims and are most suitable
for different types of signals.

Other related works include the synchrosqueezed wavelet transform \cite{Daubechies_2011_ACHA},
the iterated Hilbert transform (IHT) \cite{Gianfelici_2007_TASLP},
and more generally, algorithms for multicomponent AM-FM signal representation (see \cite{Gianfelici_2007_cnf}).
These works, again, model the instantaneous amplitudes/phase functions as smooth.

\section{Preliminaries}

\subsection{Notation}

In this paper, vectors and matrices are represented in bold (e.g. $\x$ and $\DD$).
Scalars are represented in regular font, e.g., $\Lambda$ and $\alpha$.

The $N$-point sequence $ x(n) $, $ n \in \ZZ_N = \{ 0, \dots, N-1 \} $
is denoted as the column vector 
\begin{equation}
	\x = \big[  x(0)  , \ x(1)  , \ \dots , \ x(N-1)  \big]^{t}.
\end{equation}
The $N$-point inverse DFT (IDFT) is given by
\begin{align}
	x(n) = \frac{1} {N} \sum_{k=0}^{N-1} X(k) \exp \Bigl( \myJ \frac{2 \pi k}{N} n \Bigr).
	\label{eqn:idft}
\end{align}
The IDFT can alternately be expressed in terms of sines and cosines as
\begin{align}
	x(n) = \sum_{k=0}^{N-1} \Big( a_k \cos\Bigl(  \frac{ 2 \pi k}{N} n \Bigr) + b_k \sin\Bigl( \frac{ 2 \pi k}{N} n \Bigr) \Big).
	\label{eqn:sinusoid}
\end{align}

We use $k$ to denote the discrete frequency index.
The variables $ c(n,k) $ and $ s(n,k) $ will denote $ c(n,k) = \cos(2 \pi f_k n) $ and  $ s(n,k) = \sin(2 \pi f_k n) $
for $ n \in \ZZ_N $
and
$ k \in \ZZ_K = \{0, \dots, K-1\}$.

When $ \a \in \RR^{N \times K} $ is an array, we will denote the $k$-th column by $ \a_k $,
i.e.,
\begin{equation}
	\a = [ \a_0  , \ \a_1  , \ \dots , \ \a_{K-1} ].
\end{equation}
The $\ell_1$ norm of a vector $ \v $ is denoted $ \onenorm{ \v } = \sum_i \abs{v(i) } $.
The energy of a vector $ \v $ is denoted $ \smalltwonorm{ \v }^2 = \sum_i \abs{v(i) }^2 $.

\subsection{Total Variation Denoising}

The total variation (TV) of the discrete $N$-point signal $\x$ is defined as
\begin{equation}
	\TV (\x) = \sum_{n=1}^{N-1} \ \lvert x(n) - x(n-1) \rvert .
\end{equation}
The TV of a signal is a measure of its temporal fluctuation.
It can be expressed as
\begin{equation}
	\TV (\x) = \onenorm{ \DD \x }
\end{equation}
where $ \DD $ is the matrix
\begin{equation}
	\DD = 
	\begin{bmatrix}
		-1		&1 		&		& 			\\
	   	& -1		&1		& 		&  			\\
		& 		&\ddots &\ddots	&		 	\\
		&		&		&-1		&1
	\end{bmatrix}
\end{equation}
of size $(N-1) \times N$. 
That is, $ \DD \x $ denotes the first-order difference of signal $ \x $.

Total Variation Denoising (TVD) \cite{ROF_1992} is a nonlinear filtering method defined by the 
convex optimization problem, 
\begin{align}
	\TVD ( \y  , \ \lambda ) : =  \arg \min_{\x} \smalltwonorm{ \y - \x } ^{2} + \lambda \onenorm{ \DD \x }
	\label{eqn:tvd}
\end{align}
where $\y$ is the noisy signal and $ \lam > 0 $ is a regularization parameter.
We denote the output of the TVD denoising problem by $ \TVD(\y, \lam) $ as in
\eqref{eqn:tvd}.
Specifying a large value for the regularization parameter $\lambda$ 
increases the tendency of the denoised signal to be piece-wise constant.

Efficient algorithms for 1D and multidimensional TV denoising have been developed
\cite{Chambolle_2004, Rodriguez_2009_TIP, Chartrand_2008, Osher_2005, Wang_2008_SIAM}
and applied to several linear inverse problems arising in signal and image procesing
\cite{Fig_2006_icip, DurandFroment_2003_SIAM, Steidl_2004, Wang_Zhou_2006, Oliveira_2009_SP}.
Recently, a fast and exact algorithm has been developed \cite{Condat_2012}.

\section{ Sparse Frequency Analysis}

As noted in the Introduction, we model the $N$-point signal of interest, $ \x \in \RR^N $,
as a sum of sinusoids, each with a time-varying amplitude and phase.
Equivalently, 
we can write the signal as a sum of sines and cosines, each with
a time-varying amplitude and zero phase.
Hence, we have the signal representation:
\begin{align}
	x(n)		& = \sumK \Big( a_k(n) \cos(  2 \pi f_k n ) + b_k(n) \sin( 2 \pi f_k n ) \Big)
	\label{eqn:sinusoid_2}
\end{align}
where $ f_k = k / K $ and $ a_k(n), \, b_k(n) \in \RR $, $ k \in \ZZ_K $, $ n \in \ZZ_N $.

The expansion \eqref{eqn:sinusoid_2} resembles the real form of the inverse DFT \eqref{eqn:sinusoid}.
However, in \eqref{eqn:sinusoid_2} the amplitudes $ a_k $ and $ b_k $ are time-varying.
As a consequence, there are $2NK$ coefficients in \eqref{eqn:sinusoid_2},
in contrast to $2N$ coefficients in \eqref{eqn:sinusoid}.

In addition, note that the IDFT is based on $ N $ frequencies, where $N$ is the signal length.
In contrast, the number of frequencies $ K $ in  \eqref{eqn:sinusoid_2} can be less than the signal length $N$.
Because there are $ 2KN $ independent coefficients in \eqref{eqn:sinusoid_2}, 
the signal $ \x $ can be perfectly represented with fewer than $N$ frequencies ($K < N$).

In fact, with $ K = 1 $, we can take $ a_0(n) = x(n) $ and $ b_0(n) = 0 $ and satisfy \eqref{eqn:sinusoid_2} with equality.
[With $ K = 1 $, the only term in the sum is $ a_0(n)  \cos(  2 \pi f_0 n )  $ with $ f_0 = 0 $.  ]
However, this solution is uninteresting -- the aim being to decompose
$ x(n) $ into a set of sinusoids (and cosines) with approximately piecewise constant amplitudes.

With $ K > 1 $, there are infinitely many solutions to \eqref{eqn:sinusoid_2}.
In order to obtain amplitude functions $a_k(n)$ and $ b_k(n)$ that are approximately piecewise constant, 
we regularize the total variation of $ \ak \in \RR^N $ and $ \bk \in \RR^N $ for each $ k \in \ZZ_K $,
where $ \ak $ denotes the $N$-point amplitude sequence of the $k$-th cosine component, i.e., 
 $ \a_k = (a_k(n))_{n \in \ZZ_N} $.
Similarly, $ \bk $ denotes the $N$-point amplitude sequence of the $k$-th sine component.
Regularizing $\TV(\ak)$ and $ \TV(\bk) $
promotes sparsity of the derivatives (first-order difference) of $ \ak $ and $ \bk $, as is well known.
Note that  $\TV(\ak)$ can be written as $ \norm{ \DD \ak }_1 $.
This notation will be used below.

It is not sufficient to regularize $\TV(\ak)$ and $ \TV(\bk) $ only.
Note that if $ a_k(n) $ and $ b_k(n) $ are not time-varying, i.e., $ a_k(n) = \bar{a}_k $ and $ b_k(n) = \bar{b}_k $,
then $\TV(\ak) = 0$ and $ \TV(\bk) = 0$.
Moreover, a non-time-varying solution of this form in fact exists and is given by the DFT coefficients, c.f. \eqref{eqn:sinusoid}.
This is true, provided $ K \ge N $.
That is, based on the DFT, a set of constant-amplitude sinusoids can always be found so as to perfectly represent the signal $ x(n) $.
The disadvantage of such a solution, as noted in the Introduction, is that
if $ x(n) $ contains a narrow-band component with amplitude/phase discontinuities (in addition to other components),
then many frequency components are need for the representation of the narrow-band component;
i.e., its energy is spectrally spread.
(In turn, due to each sinusoid having a constant amplitude, the energy is temporally spread as well.)
In this case, the narrow-band component can not be well isolated (extracted) from the signal $x(n)$
for the purpose of further processing or analysis (e.g., phase synchronization index estimation).

Therefore, in addition to regularizing only the total variation of the amplitudes,
we also regularize the frequency spectrum corresponding to the representation \eqref{eqn:sinusoid_2}.
We define the frequency spectrum, $ z_k $, as 
\begin{equation}
	\abs{z_k}^2 = \sum_n \, \abs{a_k(n)}^2 + \abs{b_k(n)}^2,
	\quad
	k \in \ZZ_K
\end{equation}
or equivalently
\begin{equation}
	\label{eq:zk}
	\abs{z_k}^2 = \norm{ \a_k }_2^2 + \norm{ \b_k }_2^2,
	\quad
	k \in \ZZ_K.
\end{equation}
The frequency-indexed sequence $ z_k $ measures
the distribution of amplitude energy as a function of frequency.
Note that, in case the amplitudes are not time-varying,
 i.e., $ a_k(n) = \bar{a}_k $ and $ b_k(n) = \bar{b}_k $,
 then $ \abs{z_k} $ represents the modulus of the $k$-th DFT coefficient.
We define the vector $ \z \in \RR^K$ as  $ \z = (z_k)_{k \in \ZZ_K} $. 
In order to induce $ \z $ to be sparse,
we regularize its  $\ell_1$ norm, i.e.,
\begin{equation}
	\norm{ \z }_1
	=
 	\sumK \sqrt{ \twonorm{\ak}^2 + \twonorm{\bk}^2 }.
\end{equation}

According to the preceding discussion, 
we regularize the total variation of the amplitude functions and the
$\ell_1 $ norm of the frequency spectrum
subject to the perfect reconstruction constraint.
This is expressed as the optimization problem:
\begin{multline}
	\min_{\a ,  \b} \
	\sumK \Big( \norm{\DD \ak}_{1}  + \norm{\DD \bk}_{1}
		+ 	\lambda  \sqrt{ \twonorm{\ak}^2 + \twonorm{\bk}^2 } 	 \Big) 
	\\
	\ST \ x(n)  = \sumK  \Big( \ank \, \cnk +  \bnk \,  \snk \Big)
	\tag{P0}
	\label{eqn:sfa_cost}
\end{multline}
where $ \lam > 0 $ is a regularization parameter,
and
where $\a, \b \in \mathbb{R}^{N \times K }$ with $ \a = (a(n,k))_{n \in \ZZ_N, k \in \ZZ_K} $,
and
$\a_k, \b_k \in \mathbb{R}^{N }$ with $ \a_k = (a(k,n))_{n \in \ZZ_N} $.
The $ c(n,k) $ and $ s(n,k) $ denote the cosine and sine terms,
$ c(n,k) = \cos(2 \pi f_k n) $
and
$ s(n,k) = \sin(2 \pi f_k n ) $,
where $ f_k = k/K $.

Notice in (\ref{eqn:sfa_cost}), that $\lam$ adjusts the relative weighting between the two regularization terms.
If $ K \ge 2N $,  then as $\lambda \to 0 $, the minimization problem leads to a solution approaching 
$\norm{\DD \ak}_{1} = \norm{\DD \bk}_{1} = 0$ for all frequencies $ k \in \ZZ_K $. 
In this case, $\ak$ and $\bk$ are non-time-varying.
When $K = 2N$, they coincide with the DFT coefficients,
specifically, $ K  \big( a_k(n) - \myJ b_k(n) \big) $ is equal to the DFT coefficient $X(k)$ in (\ref{eqn:idft}).

\subsection{Example 1: A Synthetic Signal}

\begin{figure}[t]
	\centering
		\includegraphics{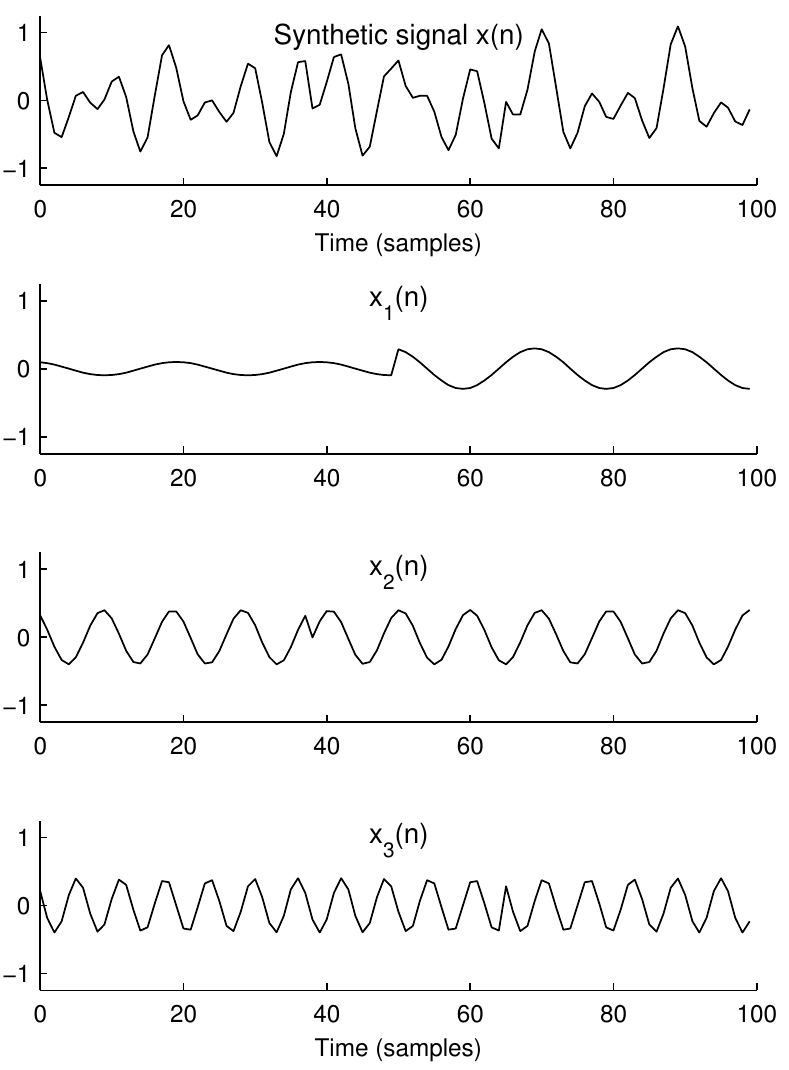} 
		\caption{
		Example 1: Test signal synthesized as the sum of three sinusoids each with a amplitude/phase discontinuity.
		}
		\label {fig:Example_test_signal}
\end{figure}

A synthetic signal $x(n)$ of length $ N = 100 $ 
is illustrated in Fig.~\ref{fig:Example_test_signal}.
The signal is formed by 
adding three sinusoidal components, $ x_i(n)$, $ i = 1,2,3$, with normalized frequencies $0.05, 0.1025$ and $0.1625$ cycles/sample, respectively.
However, each of the three sinusoidal components possess  a phase
discontinuity (`phase jump')
at $ n = 50$, 38, and 65, respectively.
The first component, $x_1(n)$, has a discontinuity in both its amplitude and phase.

\begin{figure}
	\centering
		\includegraphics{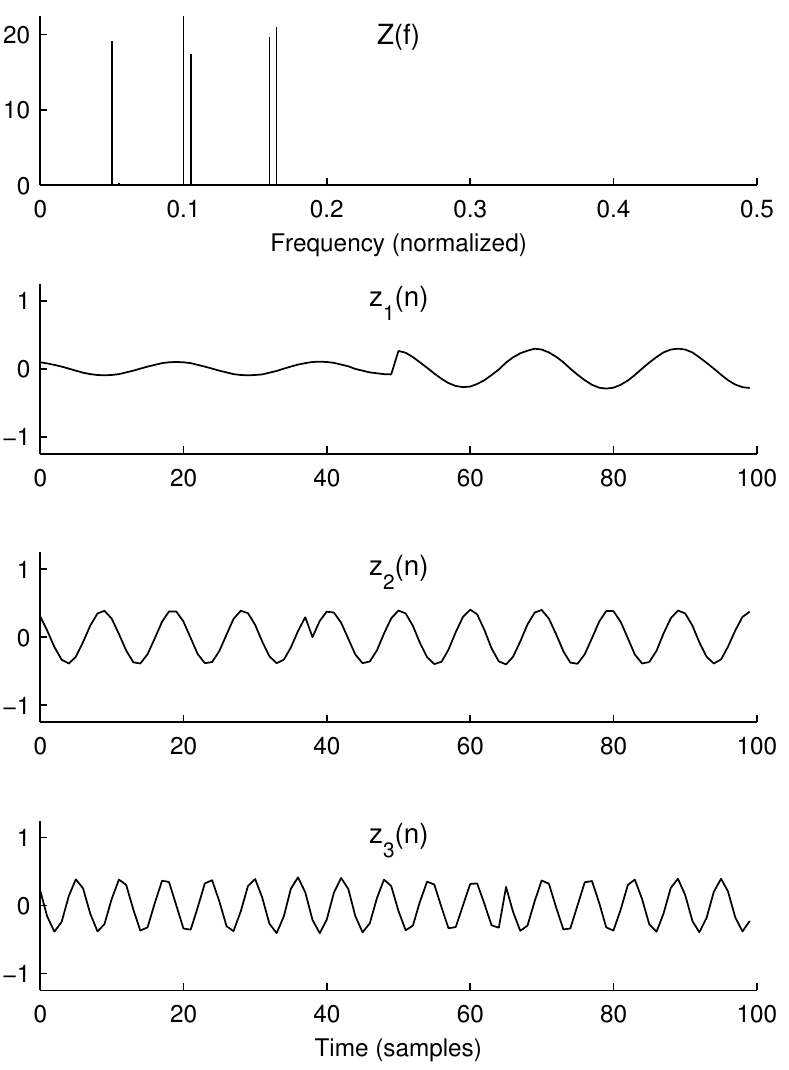}
		\caption{
		Example 1: Signal decomposition using sparse frequency analysis (SFA).
		The frequency spectrum is sparse and the recovered sinusoidal 
		components accurately retain the amplitude/phase discontinuities.
		}
		\label{fig:Example_1_SFA}
\end{figure}

\begin{figure}[!t]
	\centering
		\includegraphics{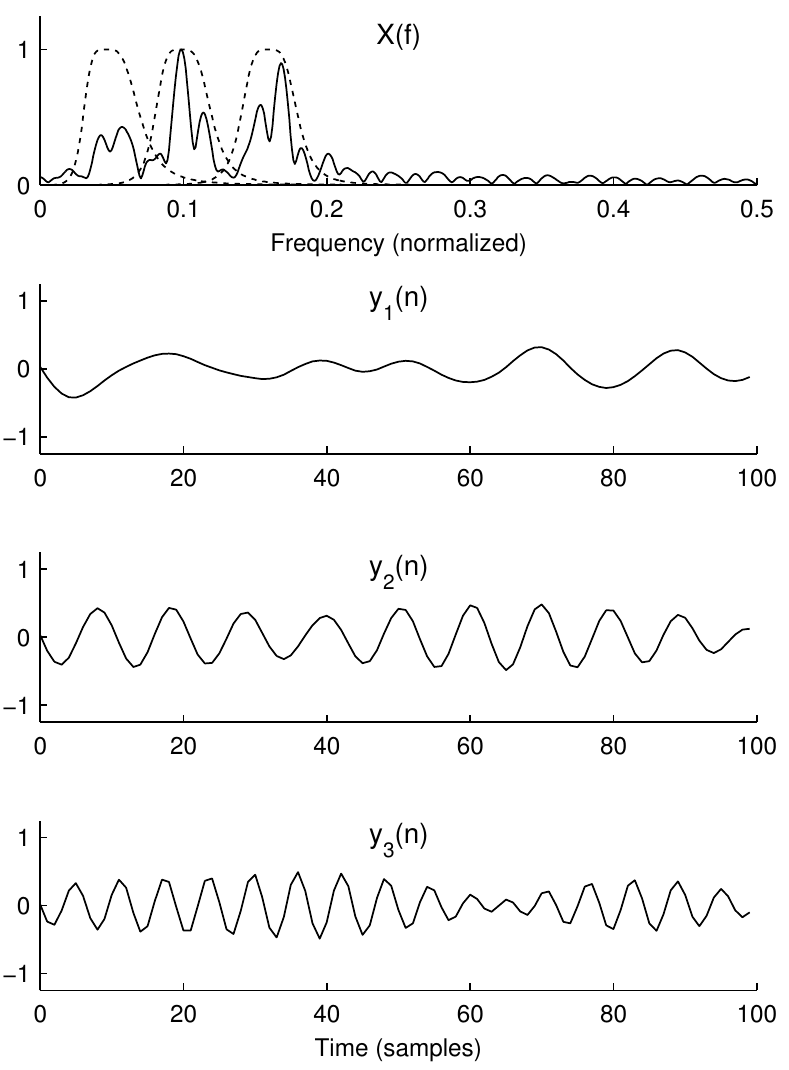}
		\caption{
		Example 1: Signal components obtained by LTI band-pass filtering. 
		The Fourier transform $X(f)$ is not sparse
		and the filtered components do not retain the amplitude/phase discontinuities. 
		The amplitude/phase discontinuities are spread across time and frequency.		
		}
		\label{fig:Example1_BPF}
\end{figure}

Here, we set $K = 100$,
so the uniformly-spaced frequencies $ f_k $, $ k \in \ZZ_K $, from 0 to 0.5,
are separated by 0.005 cycles/sample.
The frequency grid is similar to that of a 200-point DFT (including zero-padded).
The component $ x_1 $ lies exactly on the frequency grid, 
i.e. $0.05 = 10 \times 0.005 $.
On the other hand, the frequencies of components $x_2$ and $x_3$ lie exactly halfway between frequency grid points,
i.e., $ 0.1025 = 20.5 \times 0.005 $
and
$ 0.1025 = 32.5 \times 0.005 $.
The frequencies of $x_2$ and $x_3$ are chosen as such
so as to test the proposed algorithm under frequency mismatch conditions.

Using the iterative algorithm developed below, we obtain $ a(n,k) $ and $ b(n,k) $ solving the optimization problem \eqref{eqn:sfa_cost}.
The frequency spectrum, $z_k$, 
defined by \eqref{eq:zk}, is illustrated in Fig.~\ref{fig:Example_1_SFA}.
The frequency spectrum is clearly sparse.
The component $x_1$ is represented by a single line in the frequency spectrum at $ k = 10 $.
Note that, because the components $x_2$ and $x_3$ of the synthetic test signal
are not aligned with the frequency grid, they are each represented by a pair of lines in the frequency spectrum,
at $ k = (20, 21) $ and $ k = (32, 33)$, respectively.
This is similar to the leakage phenomena of the DFT, except here,
the leakage is confined to two adjacent frequency bins instead of being spread across many frequency  bins.

To extract a narrow-band signal from the proposed decomposition, defined by the arrays $ (\a, \b) $, 
we simply reconstruct the signal using a subset of frequencies,
i.e.,
\begin{equation}
	\label{eq:gS}
	g_S(n)  = \sum_{k \in S}  \Big( \ank \, \cnk +  \bnk \,  \snk \Big)
\end{equation}
where $ S \subset \ZZ_K $ is a set of one or several frequency indices.
With $ S = \{10\}$, we recover a good approximation of $ x_1 $.
With $ S = \{20, 21\}$ and $ S = \{32, 33\} $,
we recover good approximations of components $ x_2 $ and $ x_3 $, respectively.
These approximations are illustrated in
Fig.~\ref{fig:Example_1_SFA}.
Note, in particular, that the recovered components retain the
amplitude and phase discontinuities present in the original components $ x_i(n) $.
In other words,
from the signal $ x(n) $, which contains a mixture of sinusoidal components each
with amplitude/phase discontinuities, we are able to recover the components with
high accuracy, including their amplitude/phase discontinuities.

Let us compare the estimation of components $x_i$ from $x$
using the proposed method
with what can be obtained by band-pass filtering.
Band-pass filtering is widely used to analyze components of signals, e.g., 
analysis of event-related potentials (ERPs) by filtering EEG signals \cite{Kalcher_1995, Derambure_1993}.
By band-pass filtering signal $ x $ using three band-pass filters, $H_i$, $ i = 1,2,3 $,
we obtain the three output signals, $y_i$, illustrated in Fig.~\ref{fig:Example1_BPF}.
The band-pass filters are applied using forward-backward filtering to avoid phase distortion.
Clearly, the amplitude/phase discontinuities are not well preserved.
The point in each output signal, where a discontinuity is expected,
exhibits an attenuation in amplitude.
The amplitude/phase discontinuities have been spread across time and frequency.

The frequency responses of the three filters we have used
are indicated in Fig.~\ref{fig:Example1_BPF},
which also shows the Fourier transform $\abs{X(f)}$ of the signal $x(n)$.
Unlike the frequency spectrum in Fig.~\ref{fig:Example_1_SFA},
the Fourier transform in Fig.~\ref{fig:Example1_BPF} is not sparse.

\begin{figure}
	\centering
		\includegraphics{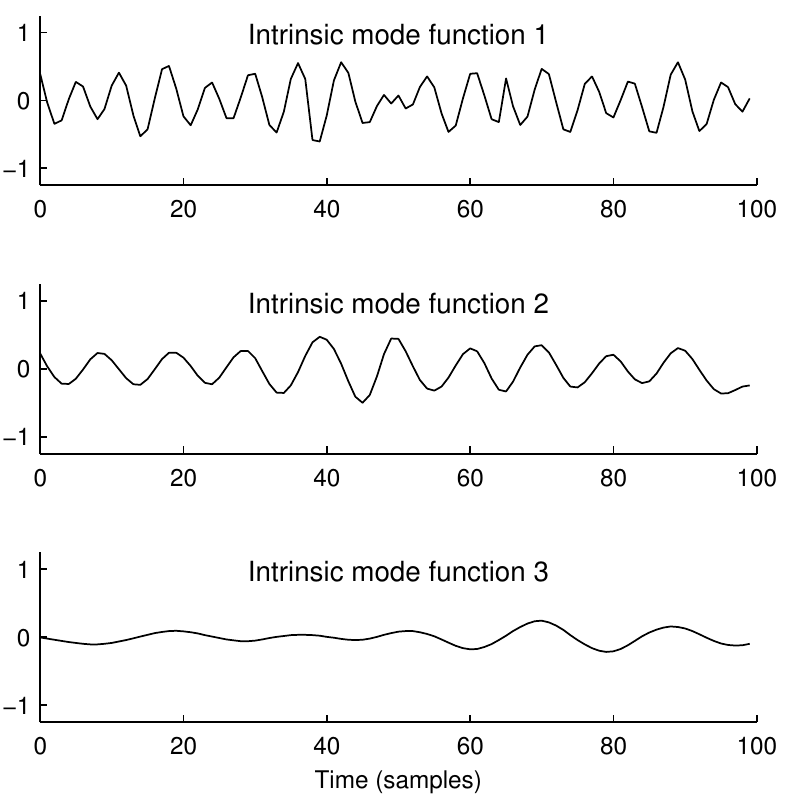}
		\caption{
		Example 1: Signal decomposition with the empirical mode decomposition (EMD).
		The first three intrinsic mode functions (IMFs).
		The amplitude/phase discontinuities are not preserved.
		}
		\label{fig:Example1_EMD}
\end{figure}

We also compare SFA and band-pass filtering with empirical mode decomposition (EMD) \cite{Huang_1998_EMD}.
EMD is a data-adaptive algorithm that decomposes a signal into a set of zero-mean oscillatory components called intrinsic
mode functions (IMFs) and a low frequency residual.
For the test signal (Fig.~\ref{fig:Example_test_signal}a), the first three IMFs 
are illustrated in Fig.~\ref{fig:Example1_EMD}.
(The EMD calculation was performed using the Matlab program \texttt{emd.m} from \url{http://perso.ens-lyon.fr/patrick.flandrin/emd.html} \cite{Rilling_2003_cnf}.)

Note that the IMFs do not capture the amplitude/phase discontinuities.
This is expected, as EMD is based on different assumptions
regarding the behavior of the narrow-band components (smooth instantaneous amplitude/phase functions).

Comparing the sparse frequency analysis (SFA) results in Fig.~\ref{fig:Example_1_SFA} 
with the band-pass filtering results in Fig.~\ref{fig:Example1_BPF}
and the EMD results in Fig.~\ref{fig:Example1_EMD},
it is clear that the SFA method is better able to extract the narrow-band components
while preserving abrupt changes in amplitude and phase.

\subsection{Optimization Algorithm}

In this section we derive an algorithm for solving problem (\ref{eqn:sfa_cost}).
We use the alternating direction method of multipliers (ADMM) and the majorization-minimization (MM) method
to transform the original problem into a sequence of simpler optimization problems,
which are solved iteratively, until convergence to the solution.

By variable splitting, problem (\ref{eqn:sfa_cost}) can be written as:
\begin {subequations}
\begin {align}	
	\min_{\a , \b , \u , \v} \	
	&	\sumK \Big( \norm{\DD \ak}_{1}  + \norm{\DD \bk}_{1} \Big) 						\nonumber 				\\
	&	\quad	+ \lambda \sumK \sqrt{ \twonorm{\uk}^2 +\twonorm{\vk}^2	} 				\nonumber				\\
	\ST \ x(n)  & =  \sumK \Big( \unk \, \cnk + \vnk \, \snk \Big)					\label{eqn:sfa_cost_2_b}	\\
	\u  & = \a 			
	\\
	\v 	& = \b 																			
\end {align}
\end {subequations}
where $\u , \v \in \mathbb{R}^{ N \times K }$ correspond to matrices $\a$ and $\b$.

The augmented Lagrangian \cite{Afonso_2010_TIP_SALSA}
is given by
\begin{align}
	&	L_0(\a , \b, \u, \v , \lambda , \mu ) 											\nonumber 				\\
	&	\quad	=	\sumK \bigg(  \onenorm{\DD \ak} + \onenorm{\DD \bk}	
				+ 	\lambda \sqrt{ \twonorm{\uk}^2 + \twonorm{\vk}^2 }	\ 			\nonumber 				\\[0.4em]
	& 	\quad	\quad 	+ 	\mu  	\twonorm{\uk - \ak - \pk }^{2}	
						+ 	\mu \twonorm{\vk - \bk - \qk }^{2} \bigg) 	\label{eqn:sfa_AL}					
\end{align}
where $\mu > 0$ is a parameter to be specified,
and where the equality constraint (\ref{eqn:sfa_cost_2_b}) still holds.
The parameter $\mu$ does not influence the minimizer of cost function (\ref{eqn:sfa_cost}).
Therefore, the solution of (\ref{eqn:sfa_AL}) leads to the solution of (\ref{eqn:sfa_cost}).
The use of ADMM yields the iterative algorithm:
\begin{subequations}																		\label{eqn:sfa_ADMM_2}
\begin{align}
	& \a , \b \gets \,
	 		\arg \min_{\a,\b}  \sumK \Big( \onenorm{\DD \ak} + \onenorm{\DD \bk}	\Big.	\nonumber				\\
	&	\qquad 	+ 	\Big. 	\mu \twonorm{\uk - \ak - \pk }^{2}	
				+ 			\mu \twonorm{\vk - \bk - \qk }^{2}	\Big)					\label{eqn:sfa_ADMM_2_a} \\
	& \u, \v \gets 	\,
	 		\arg \min_{\u, \v} 	
			 \sumK \Big( \lambda \sqrt{ \twonorm{\uk}^2 + \twonorm{\vk}^2 }				\nonumber				\\			
	&	\qquad 	+ 	\mu \twonorm{\uk - \ak - \pk }^{2}	
				+ 	\mu \twonorm{\vk - \bk - \qk }^{2}	\Big)							\nonumber				\\
	& \ST x(n)   =  \sumK \Big( \unk \, \cnk + \vnk \, \snk \Big)					\label{eqn:sfa_ADMM_2_b}	\\
	& \p  \gets \  \p - (  \u - \a  ) 																			\\
	& \q  \gets \  \q - (  \v - \b  )  																			\\
	&	\text{Go back to (\ref{eqn:sfa_ADMM_2_a}).} 
\end{align}
\end{subequations}

In (\ref{eqn:sfa_ADMM_2_a}), the variables $\a$ and $\b$ are uncoupled.
Furthermore, each of the $K$ columns $\ak$ and $\bk$ of $\a$ and $\b$ are decoupled.
Hence, we can write
\begin{subequations}																		\label{eqn:sfa_ADMM_ab}
\begin{align}
	\ak  \gets \,
	& 	\arg \min_{\ak} \  \onenorm{\DD \ak} +	\mu \twonorm{\uk - \ak - \pk }^{2}								\\[0.4em]
	\bk 	\gets \,
	& 	\arg \min_{\bk} \  \onenorm{\DD \bk} + 	\mu \twonorm{\vk - \bk - \qk }^{2}
\end{align}	
\end{subequations} 
for $ k \in \ZZ_K $.
Problems \eqref{eqn:sfa_ADMM_ab} are recognized as $N$-point TV denoising problems,
readily solved by \cite{Condat_2012}.
Hence we write
\begin{subequations}																		\label{eqn:sfa_ab}
\begin{align}
	\ak \gets \ 	& 	\TVD( \uk - \pk  , \,  1/\mu )														\\[0.4em]	
	\bk \gets \ 	& 	\TVD( \vk - \qk  , \,  1/\mu )
\end{align}	
\end{subequations} 
for $ k \in \ZZ_K $.

Problem \eqref{eqn:sfa_ADMM_2_b} can be written as:
\begin{multline*}
 	\u , \v  
	 \gets \
		\arg \min_{\u , \v}
		\sumK \Bigg[ \lam \sqrt{ \sumN  \vert \unk \vert^{2} +  \vert \vnk \vert^{2}  }	
	\\
	+ \mu \   \sumN | \unk - \ank - \pnk | ^ {2} 	
	\\
	+ \mu \   \sumN | \vnk - \bnk - \qnk | ^ {2} 	\ \Bigg]
\end{multline*}
\begin{equation}
	\label{eqn:sfa_u_2}
	\ST x(n)  =  \sumK \Big( \unk \, \cnk + \vnk \, \snk \Big)						
\end{equation}
which does not admit an explicit solution.
Here we use the MM procedure for solving \eqref{eqn:sfa_u_2}, i.e., \eqref{eqn:sfa_ADMM_2_b}.
First we need a majorizer. 
To that end, note that for each $k \in \ZZ_K$,
\begin{multline}
\label{eqn:sfa_majorizer_a}
	\sqrt{ \sumN  \vert \unk \vert^{2} +  \vert \vnk \vert^{2}  }
	\\
	\le \frac{1}{2 \Lambda_k\iter{i}} \sumN \big( \vert \unk \vert^{2} +  \vert \vnk \vert^{2} \big) + \frac{1}{2}\Lambda_k\iter{i}
\end{multline}
where
\begin{equation}
	\Lambda_k\iter{i} = \sqrt{ \sumN \vert \uink \vert^{2} +  \vert \vink \vert^{2} }.
	\label{eqn:sfa_majorizer_lam}	
\end{equation}
Here $i$ is the iteration index for the MM procedure
and the right-hand side of \eqref{eqn:sfa_majorizer_a} is the majorizer.
An MM algorithm for solving  (\ref{eqn:sfa_u_2}) is hence given by:
\begin{align}
 	& \u^{(i+1)} , \ \v^{(i+1)} 																	\nonumber		\\
	&	\gets \ \arg \min_{\u , \v}  
		\sumK  \Bigg[ \  \sumN \biggl( \frac{\lambda}{2 \Lambda_k\iter{i}}  \big( \vert \unk \vert^{2} +  \vert \vnk \vert^{2} \big) 
																								\nonumber		\\[0.4em]
	&	\qquad \ \qquad \ + \mu \,	\abs{ \unk - \ank - \pnk }^2								\nonumber		\\
	&	\qquad \ \qquad \ + \mu \,	\abs{ \vnk - \bnk - \qnk }^2 \biggr)  \Bigg]						\nonumber		\\
	& \ST x(n)  =  \sumK \Big( \unk \, \cnk + \vnk \, \snk \Big).								\label{eqn:sfa_u_3}										
\end{align}
The majorizer is significantly simpler to minimize.
With the use of the quadratic term to majorize the $\ell_2$-norm, 
the problem becomes separable with respect to $n$.
That is, \eqref{eqn:sfa_u_3} constitutes $ N $ independent
least-square optimization problems.
Moreover, each least-square problem is relatively simple and admits
an explicit solution.
Omitting straightforward details,
$\u\iter{i+1}$ and $\v\iter{i+1}$ solving \eqref{eqn:sfa_u_3} are given explicitly by:
\begin{subequations}
\label{eqn:sfa_u_5}
\begin{align}
	u\iter{i+1}(n,k) 	& = 	V_k \Big[ \alpha(n,k) \, c(n,k) 												\nonumber		\\
			&	\qquad \qquad \qquad + 2 \mu \big( \ank + \pnk \big) \Big]  									\\
	v\iter{i+1}(n,k) 	& =	V_k \Big[ \alpha(n,k) \, s(n,k) 												\nonumber		\\
			&	\qquad \qquad \qquad + 2 \mu \big( \bnk + \qnk \big) \Big]
\end{align}
\end{subequations}
where
\[
	\alpha(n,k) = \Big[\sumK V_k \Big]^{-1} \Big[ x(n) - 2 \mu \sumK \gamma(n,k) \Big]
\]
\begin{multline}
	\gamma(n,k) = V_k \Big[ c(n,k) \big( \ank + \pnk \big)  					
	\\	
		 + s(n,k) \big( \bnk + \qnk \big) \Big]
\end{multline}
and
\begin{equation}
	\label{eqn:sfa_Vk}
	V_k = \frac { \Lambda_k\iter{i} }{  2 \mu  \Lambda_k\iter{i} + \lambda  }
\end{equation}
for $ n \in \ZZ_N $, $ k \in \ZZ_K $.
Equations \eqref{eqn:sfa_majorizer_lam} and \eqref{eqn:sfa_u_5} constitute an MM algorithm
for computing the solution to  \eqref{eqn:sfa_ADMM_2_b}.
Numerous MM iterations are required in order to obtain an accurate solution to \eqref{eqn:sfa_ADMM_2_b}.
However, due to the nesting of the MM algorithm within the ADMM algorithm,
it is not necessary to perform many MM iterations for each ADMM iteration.

\begin{figure}
	\centering
		\includegraphics[scale = \figurescale] {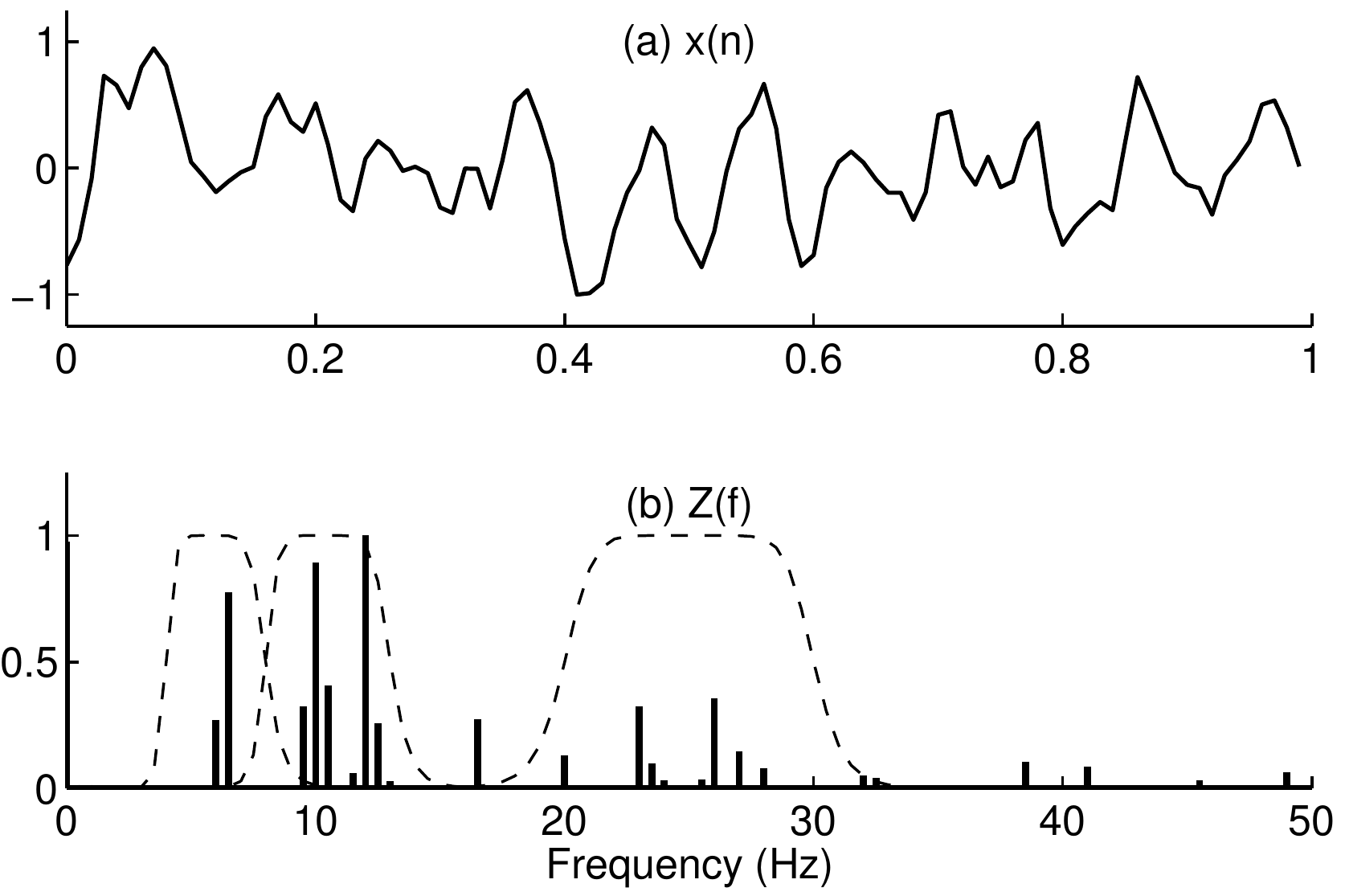}
		\includegraphics[scale = \figurescale] {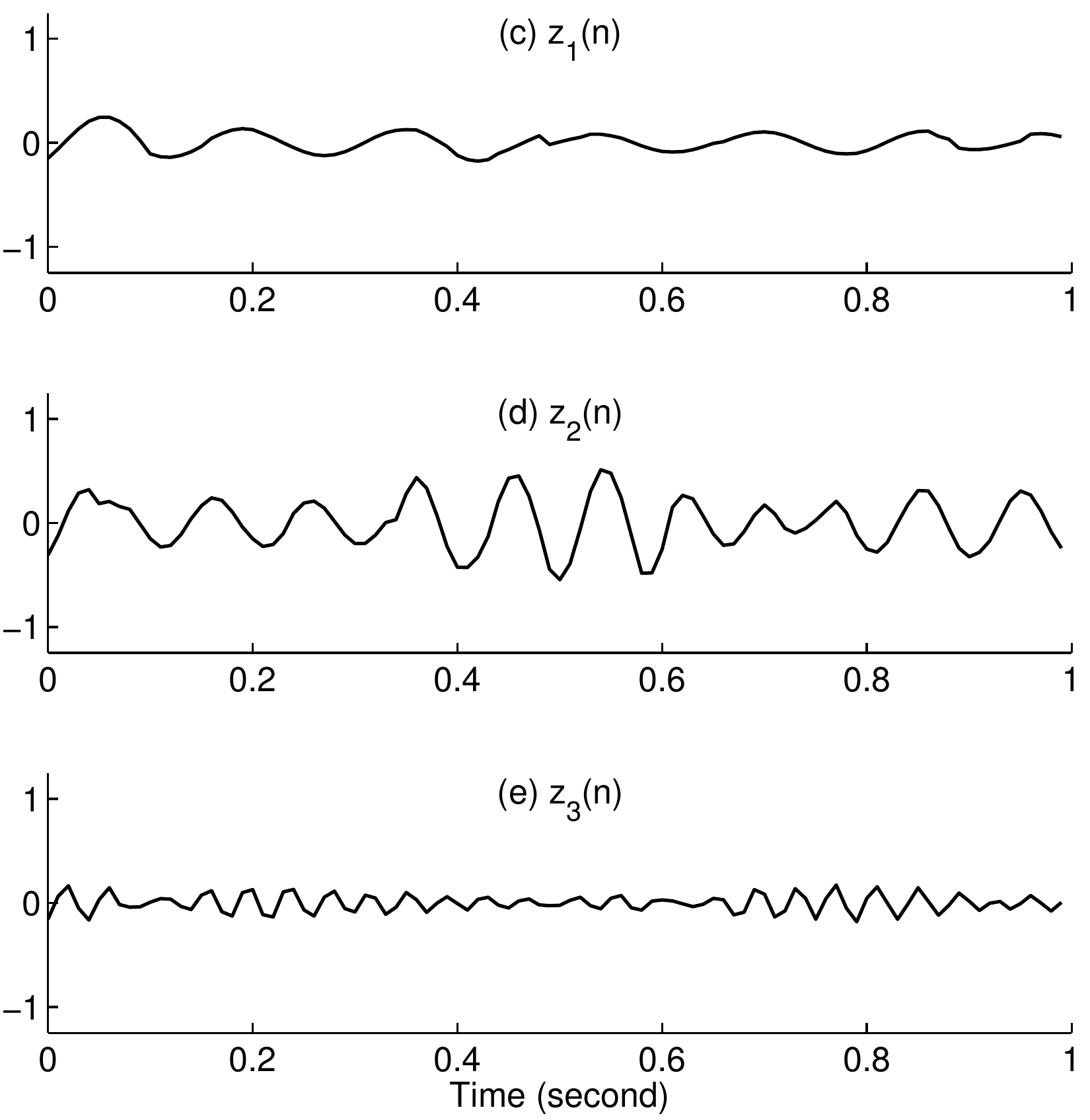}
		\caption{
		Example 2:
		EEG signal, sparse frequency spectrum obtained using sparse frequency analysis (SFA), 
		band-pass components reconstructed from SFA decomposition.
		}
		\label{fig:Example2_SFA}
\end{figure}

\begin{figure}
	\centering
		\includegraphics[scale = \figurescale] {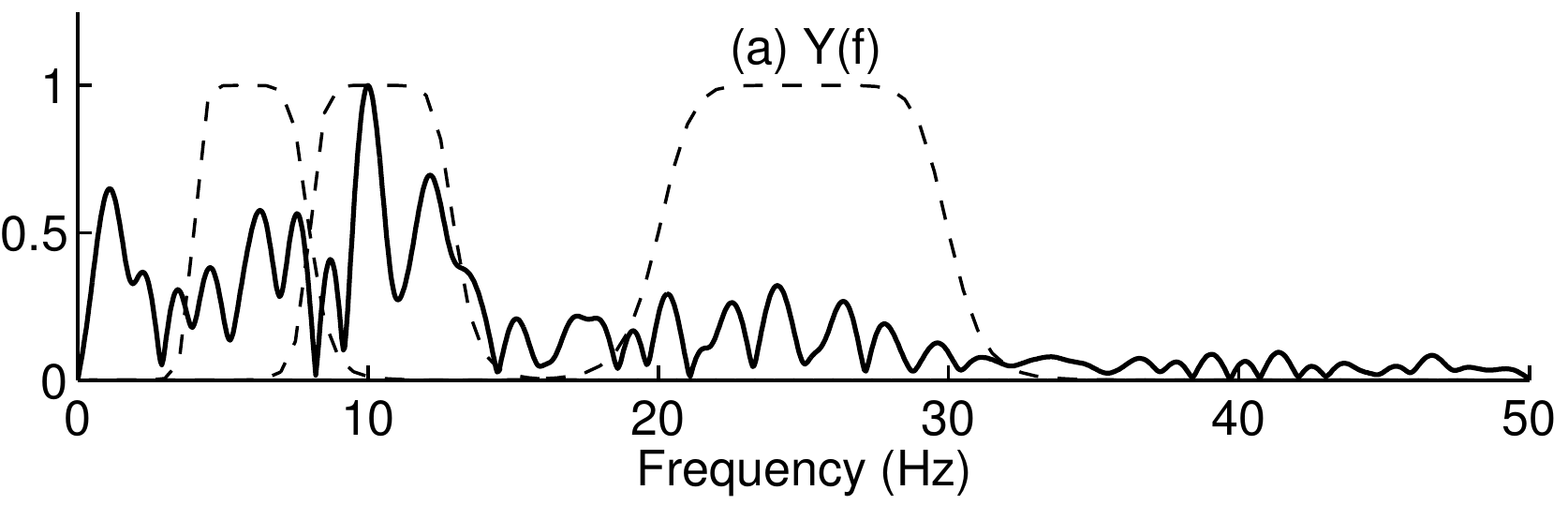}
		\includegraphics[scale = \figurescale] {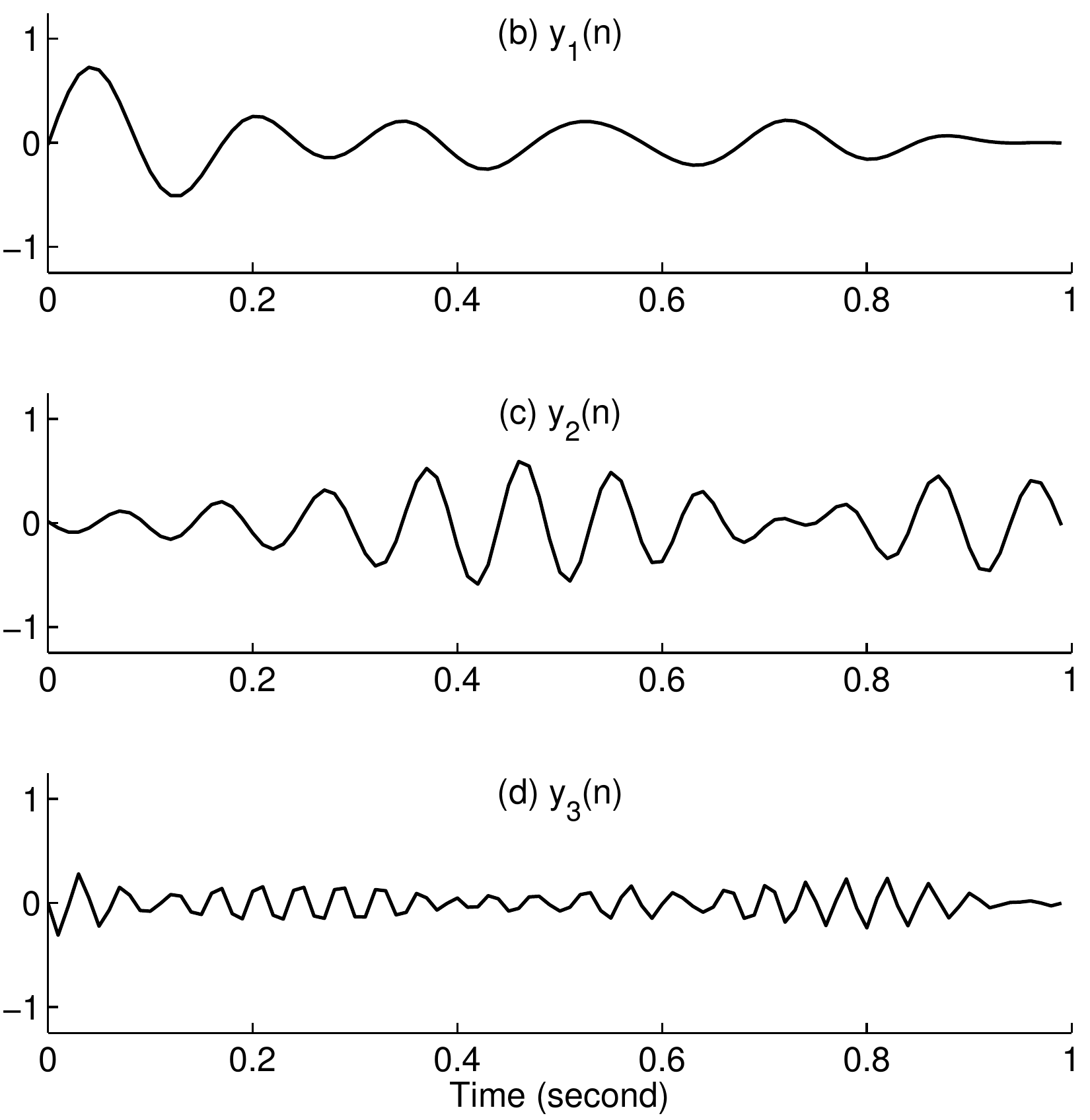}
		\caption{
		Example 2:
		Fourier transform of EE signal 
		and band-pass components obtained using LTI band-pass filtering.
		}
		\label{fig:Example2_BPF}
\end{figure}

\subsection{Example 2: EEG Signal}

An electroencephalogram (EEG) is the measurement of electrical activity of the brain.
Several frequency bands have been recognized as having physiological significance, 
for example:
$4 \le f \le 8 $ Hz (theta rhythms),
$8 \le f \le 13 $ Hz (alpha rhythms),
$ 13 \le f \le 30 $ Hz (beta rhythms).
These rhythms are usually obtained by band-pass filtering EEG data \cite{Rangayyan_book}.

A one-second EEG signal from \cite{Rangayyan_book}, with sampling rate $f_s = 100$ Hz,
is illustrated in Fig.~\ref{fig:Example2_SFA}.
In this example,
we aim to estimate the three noted rhythms
via the proposed sparse frequency analysis method, and compare the result with that of band-pass filtering.

The proposed method yields the sparse frequency spectrum illustrated in Fig.~\ref{fig:Example2_SFA}.
In order to implement (non-linear) band-pass filtering using the proposed method,
we simply reconstruct the signal by weighting each frequency component by the
frequency response of a specified band-pass filter,
\[
	g_H(n)  = \sum_{k \in \ZZ_K} \abs{H(f_k)} \Big( \ank \, \cnk +  \bnk \,  \snk \Big).
\]
This generalization of \eqref{eq:gS} 
incorporates the frequency response of the band-pass filter, $H$.

Three band-pass filters $H_i$, $i = 1, 2, 3$, are designed according to the theta, alpha, and beta rhythms.
Their frequency responses are illustrated in Fig.~\ref{fig:Example2_SFA},
overlaid on the sparse spectrum obtained by the proposed technique.
The three reconstructed components $g_{H_i}$ are illustrated in Fig.~\ref{fig:Example2_SFA}.
It can be seen that each component has relatively piecewise-constant amplitudes and phases.
For example, the component $ g_{H_1} $ in Fig.~\ref{fig:Example2_SFA}(c)
exhibits an amplitude/phase discontinuity at about $ t = 0.5 $ seconds.
The component $ g_{H_2} $ in Fig.~\ref{fig:Example2_SFA}(d)
exhibits an amplitude/phase discontinuity at about $ t = 0.35 $ seconds
and shortly after $ t = 0.6 $ seconds.
The (instantaneous) amplitude/phase functions are otherwise relatively constant.

The signals obtained by LTI band-pass filtering are shown in Fig.~\ref{fig:Example2_BPF}.
The utilized band-pass filters are those that were used for the sparse frequency approach.
The Fourier transform of the EEG signal is shown in Fig.~\ref{fig:Example2_BPF}(a).

Comparing the estimated theta, alpha, and beta rhythms
obtained by the two methods, shown in Figs.~\ref{fig:Example2_SFA} and \ref{fig:Example2_BPF},
it can be seen that they are quite similar.
Hence, the proposed method gives a result that is reasonably similar to LTI band-pass filtering, as desired.
Yet, the proposed approach provides a potentially more accurate
estimation of abrupt changes in amplitude and phase.
In this example, the true components are, of course, unknown.
However, the sparse frequency analysis approach provides an alternative to LTI filtering,
useful in particular, where it is thought the underlying components are
sparse in frequency and possess sparse amplitude and phase deriviatives.

\section{Sparse Frequency Approximation}

In applications, the available data $ y(n) $ is usually somewhat noisy
and it may be unnecessary or undesirable to enforce the perfect reconstruction constraint in \eqref{eqn:sfa_cost}.
Here we assume the data is of the form $\y = \x + \w$, where $\w$ denotes additive white Gaussian noise.
In this case, a problem formulation more suitable than \eqref{eqn:sfa_cost}
is the following one, where, as in basis pursuit denoising (BPD) \cite{Chen_1998_SIAM},
an approximate representation of the data is sought.
\begin{multline}
	\min_{\a , \b} \
	\sumK \Big( \norm{\DD \ak}_{1}  + \norm{\DD \bk}_{1} 
		+ \lambda  \sqrt{ 	\twonorm{\ak}^2 + \twonorm{\bk}^2	} \Big)										\\
	+ \lambda_1 \sumN \bigg[ \ y(n) - \sumK \Big( \ank \, \cnk + \bnk \, \snk \Big) \bigg]^{2} 
	\tag{P1}
	\label{eqn:sfad_cost}
\end{multline}
The parameter $ \lam_1 > 0 $ should be chosen according to the noise level.
In the following, we derive an algorithm for solving \eqref{eqn:sfad_cost}.

Applying variable splitting, as above, \eqref{eqn:sfad_cost} can be rewritten:
\begin{multline}
	\nonumber
	 \min_{\a , \b , \u , \v} 
		\sumK \Big( \norm{\DD \ak}_{1}  + \norm{\DD \bk}_{1} + \lambda \sqrt{ 	\twonorm{\uk}^2 + \twonorm{\vk}^2	}  \Big)			
	\\[0.2em]
		+ \lambda_1 \sumN \bigg[ \ y(n) - \sumK \Big( \unk \, \cnk + \vnk \, \snk \Big) \bigg]^{2}				
	\\
	\shoveleft
	 \ST  %
	\left\{ \begin{aligned} & \u = \a \\  & \v = \b. \end{aligned}	\right.
\end{multline}
As above, we apply ADMM, to obtain the iterative algorithm:
\begin{subequations}																			\label{eqn:sfad_ADMM}
\begin{align}
	& \a , \b \gets \,
	 	\arg \min_{\a,\b}  \sumK \Big( \onenorm{\DD \ak} + \onenorm{\DD \bk}					\nonumber				\\
	&	\quad +\mu \twonorm{\uk - \ak - \pk }^{2} + \mu \twonorm{\vk - \bk - \qk }^{2} \Big)	\label{eqn:sfad_ADMM_a}  	\\[0.8em] 
	& \u, \v \gets 	\,
																							\label{eqn:sfad_ADMM_b}
	 		\arg \min_{\u, \v} 	
			\sumK \Big( \lambda \sqrt{ \twonorm{\uk}^2 + \twonorm{\vk}^2 }								\\[0.4em]	
	&	\quad 	+ \mu \twonorm{\uk - \ak - \pk }^{2} + \mu \twonorm{\vk - \bk - \qk }^{2}\Big)		\nonumber			\\[0.4em]
	&		+	\lambda_1 \sumN \! \bigg[  y(n)  - \!
	 \sumK \! \! \Big( \! \unk \cnk +  \vnk \snk \! \Big)	\! \bigg]^{2}	
																			\nonumber					\\[0.8em]
	& \p  \gets \,  \ \p - ( \u - \a ) 																			\\[0.4em]
	& \q  \gets \,  \ \q - ( \v - \b )  																			\\[0.8em]
	&	\text{Go to \eqref{eqn:sfad_ADMM_a}.} 			
\end{align}
\end{subequations} 
Note that step (\ref{eqn:sfad_ADMM_a}) is exactly the same as (\ref{eqn:sfa_ADMM_2_a}),
the solution of which is given by (\ref{eqn:sfa_ab}), i.e., TV denoising.
To solve \eqref{eqn:sfad_ADMM_b} for $\u$ and $\v$, 
we use MM with the majorizer given in \eqref{eqn:sfa_majorizer_a}. 
With this majorizor, an MM algorithm to solve \eqref{eqn:sfad_ADMM_b}
is given by the following iteration, where $i$ is the MM iteration index.
\begin{align}
\label{eqn:sfad_u_1}
 	&\u^{(i+1)} , \v^{(i+1)}  																	\nonumber			\\
	&	\gets \ \arg \min_{\u , \v}   
		\sumN \Bigg[ \sumK \frac{\lambda}{2 \Lambda_k\iter{i}} \big( \vert \unk \vert^{2} +  \vert \vnk \vert^{2} \big)	
																								\nonumber			 	\\
	&	\ +	\lambda_1 \bigg( y(n)  - \sumK \big( \unk \, \cnk +  \vnk \, \snk \big)	 \bigg)	^{2}			
																								\nonumber 				\\
	&	\ + \mu   \sumK | \unk - \ank - \pnk | ^ {2} 								\nonumber				\\
	&	\ + \mu   \sumK | \vnk - \bnk - \qnk | ^ {2} 	\ \Bigg]			
\end{align}
where $ \Lambda_k\iter{i} $ is given by \eqref{eqn:sfa_majorizer_lam}.
As in \eqref{eqn:sfa_u_3}, problem \eqref{eqn:sfad_u_1} decouples with respect to $n$
and the solution $\u\iter{i+1}$, $\v\iter{i+1}$ can be found in explicit form:
%
%
\begin{align*}
	u\iter{i+1}(n,k) 	& = 	V_k \Big[ \beta(n,k) \, c(n,k) + 2 \mu \big( \ank + \pnk \big) \Big] 
	\\
	v\iter{i+1}(n,k) 	& =	V_k \Big[ \beta(n,k) \, s(n,k) + 2 \mu \big( \bnk + \qnk \big) \Big]								
\end{align*}
where
\[
	\beta(n,k) = \Big[ \frac{1}{2\lambda_1}+ \sumK V_k \Big]^{-1}
					\Big[ y(n) - 2\mu \sumK \gamma(n,k) \Big]
\]
\begin{multline}
	\gamma(n,k) 	= V_k \Big[ c(n,k) \big( \ank + \pnk \big) 
	\\
			 + s(n,k) \big( \bnk + \qnk \big) \Big]
\end{multline}
where $V_k$ is given by \eqref{eqn:sfa_Vk}.

\subsection{Example 3: A Noisy Multiband Signal}

This example illustrates the estimation of a sinusoid with a phase discontinuity 
when the observed data includes numerous other components, including additive white Gaussian noise.
The example also illustrates the estimation of the instantaneous phase
of the estimated component. 
The resulting instantaneous phase function is compared with that obtained using a band-pass filter and the Hilbert transform.
The Hilbert transform is widely used in applications such as EEG analsys \cite{Ktonas_1980, Medl_1992, Vairavan_2009, Wacker_2011}.

\begin{figure}
	\centering
		\includegraphics[scale = \figurescale]{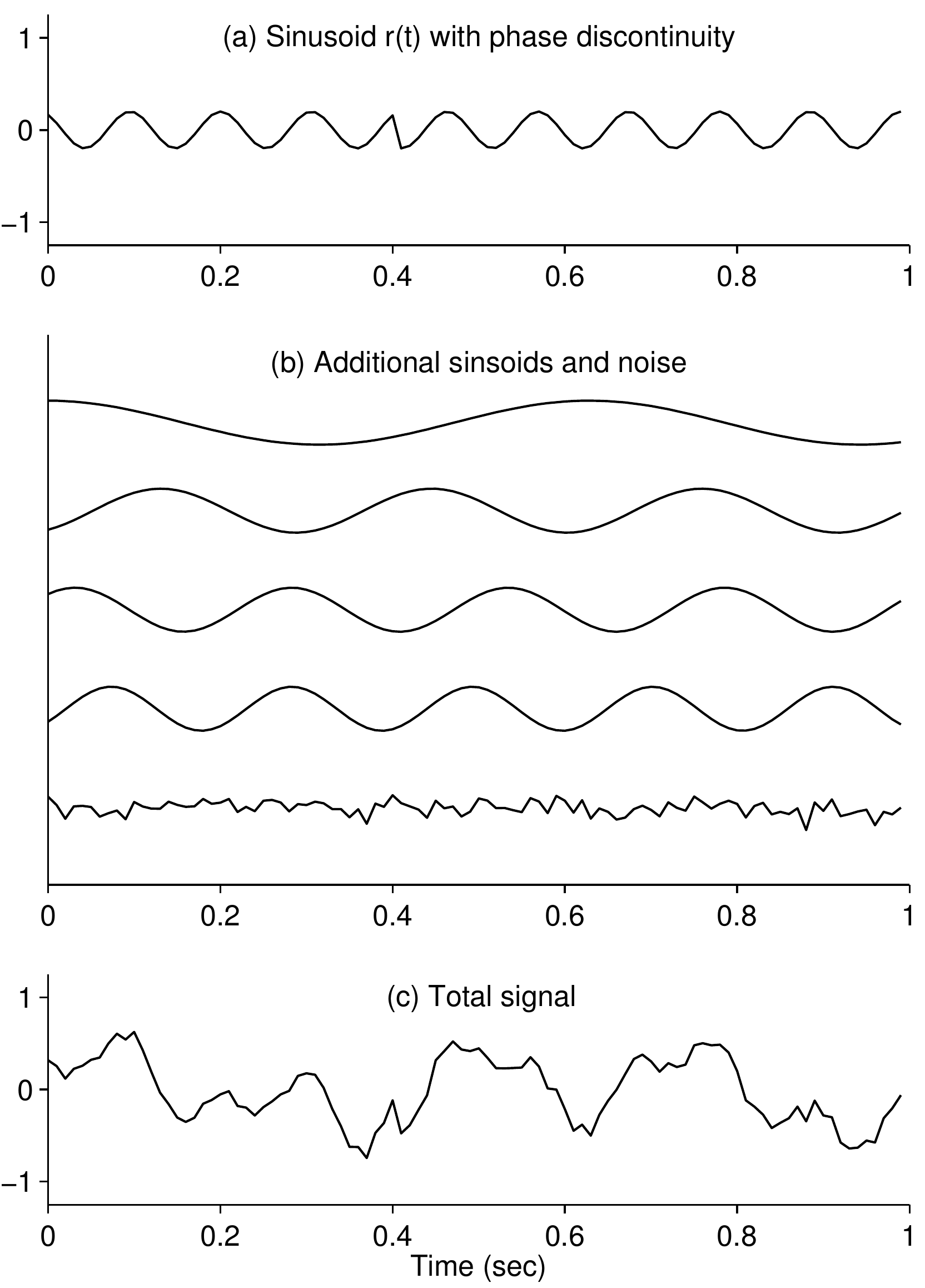}
		\caption{
		Example 3:
		The signal $x(t)$ in (c) is synthesized as the sum of 
		(a) 9.5 Hz sinusoid with a phase discontinuity and (b)
		additional sinusoids and white Gaussian noise.
		}
		\label{fig:Example_3_test_signal}
\end{figure}

\begin{figure}[!t]
	\centering
		\includegraphics[scale = \figurescale]{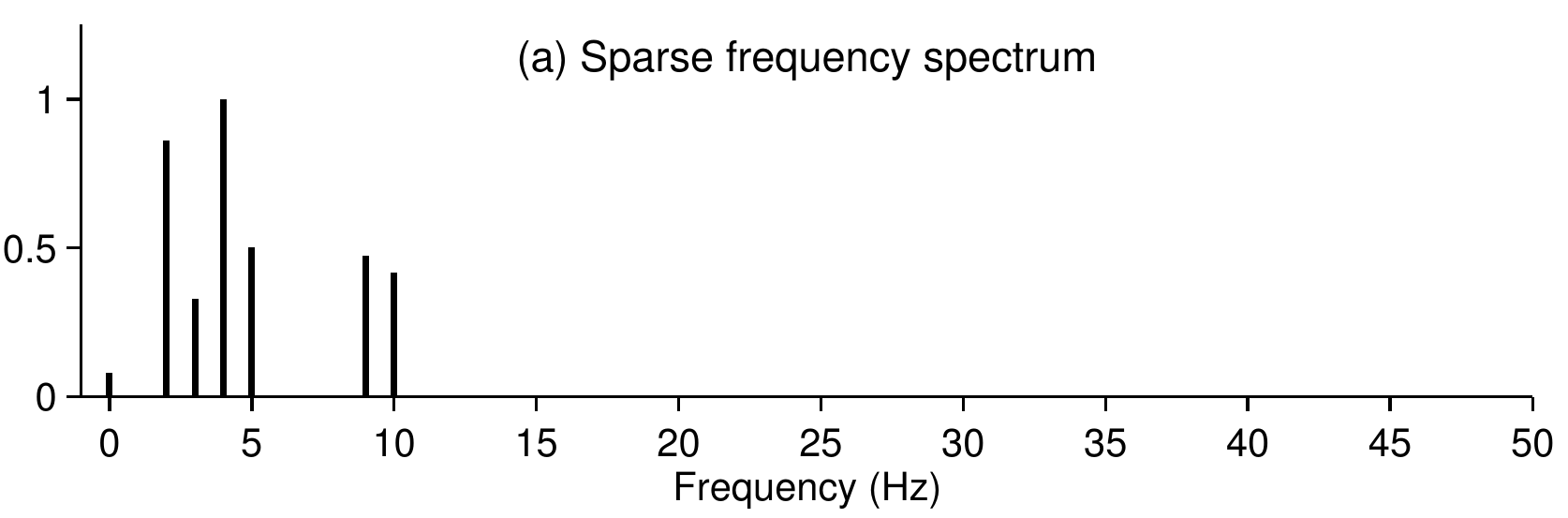} 
		\smallskip
		\includegraphics[scale = \figurescale]{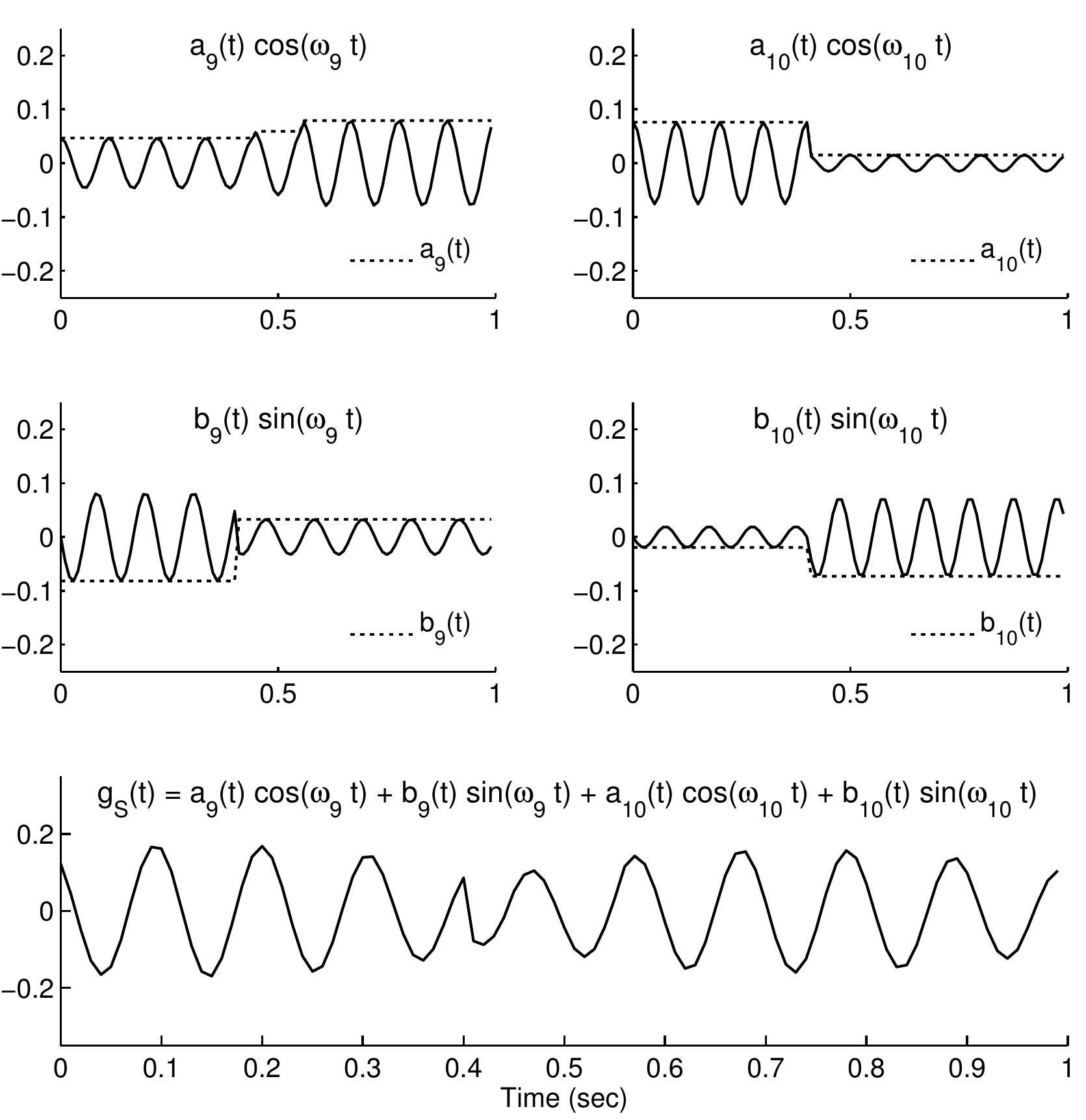} 
		\smallskip
		\includegraphics[scale = \figurescale]{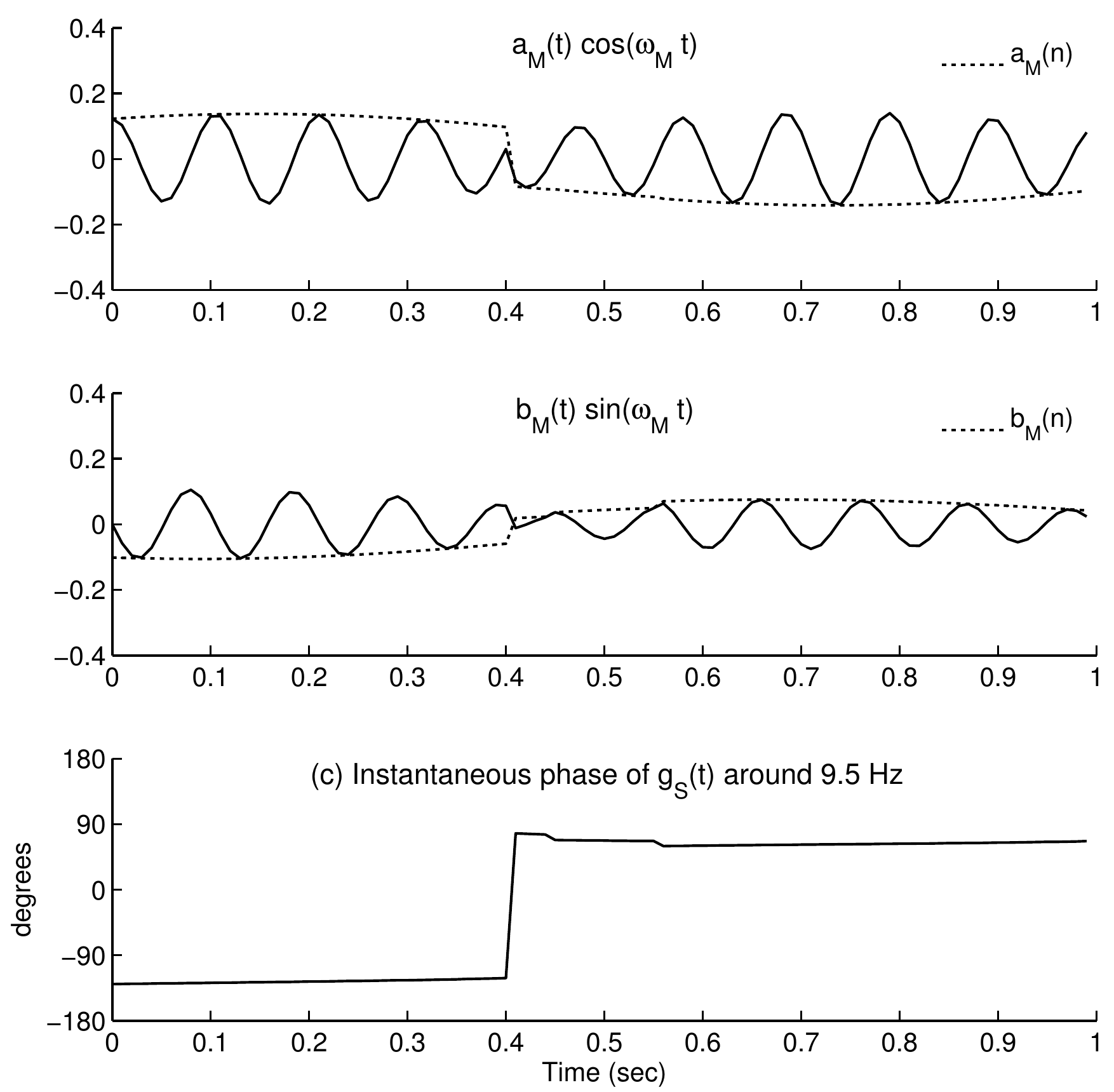} 
		\caption{
		Example 3: Signal decomposition using sparse frequency analysis (SFA).
		The discontinuity in the instantaneous phase of the 9.5 Hz sinusoid is accurately recovered.		
		}
		\label{fig:Example_artifical_sfad_recover}
\end{figure}

The one-second signal $ x(t) $, with sampling rate of $ F_s = 100 $ Hz ($ N = 100 $ samples),
illustrated in Fig.~\ref{fig:Example_3_test_signal}c,  is synthesized as follows:
adding the 9.5 Hz sinusoid $r(t)$ (with phase discontinuity at $ t = 0.4 $ seconds) in Fig.~\ref{fig:Example_3_test_signal}a,
and the sinusoids and white Gaussian noise illustrated in Fig.~\ref{fig:Example_3_test_signal}b.
Note that the instantaneous frequency of $ r(t) $ has an impulse at $ t = 0.4 $ due to the phase discontinuity.

The sparse frequency analysis approach \eqref{eqn:sfad_cost} was solved
with the iterative algorithm described above.
We used
$ K = 50 $ uniformly spaced frequencies from 0 to 50 Hz,
we obtain $ \a, \b \in \RR^{N\times K} $,
the time-varying cosine and sine amplitudes,
with discrete frequencies $ f_k = k $ Hz.
The sparse frequency spectrum is illustrated in Fig.~\ref{fig:Example_artifical_sfad_recover}(a).
Our aim is to recover $ r(t) $, but (by design)
its frequency of 9.5 Hz is halfway between the discrete frequencies $f_9$ and $ f_{10} $,
which are clearly visible in the sparse spectrum.
Hence, an estimate of $r(t)$ is obtained via \eqref{eq:gS} using the index set $ S = \{ 9, 10 \} $,
i.e., $ \hat{r}(t) = g_{\{9, 10\}}(t) $,
\begin{multline}
	g_S(t) = 
	a_9(t) \cos\om_9 t + b_9(t) \sin\om_9 t
	\\
	+ a_{10}(t) \cos\om_{10} t + b_{10}(t) \sin\om_{10} t .
\end{multline}
The time-varying amplitudes, $ a_k(t) $ and $ b_k(t) $ for $ k = 9, 10 $, are
illustrated in Fig.~\ref{fig:Example_artifical_sfad_recover} (dashed lines).
Within the same plots, the functions $ a_k(t) \cos(\om_k t) $ and $ b_k(t) \sin(\om_k t) $ with $ \om_k  = 2\pi f_k $ for $ k = 9, 10 $ are also shown (solid lines).
The piecewise-constant property of $ a_k(t) $ and $ b_k(t) $ is clearly visible.
The signal $g_S(t) $ is also illustrated in the figure.
Note that $ g_S(t) $ has a center frequency of 9.5 Hz, while the
sinusoids from which it is composed have frequencies 9 and 10 Hz.

To compute and analyze the instantaneous phase of the signal $ g_S(t) $,
it is convenient to express $g_S(t)$ in terms of a single frequency (9.5 Hz) instead of two distinct frequencies (9 and 10 Hz).
Therefore, by trigonometric identities, we write
\begin{multline}
	\label{eq:gST}
	g_S(t) = 
	a_m(t) \cos(\om_m t) + b_m(t) \sin(\om_m t)
	\\
	+
	a_{m+1}(t) \cos(\om_{m+1} t) + b_{m+1}(t) \sin(\om_{m+1} t)
\end{multline}
as
\begin{equation}
	\label{eq:gSM}
	g_S(t) = 
	a_M(t) \cos(\om_M t) + b_M(t) \sin(\om_M t)
\end{equation}
where
\begin{multline}
	\label{eq:aM}
	a_M(t) = 
		\big( a_{m}(t) + a_{m+1}(t) \big) \cos( \dw t ) 				
	\\
	 - \big( b_{m}(t) - b_{m+1}(t) \big) \sin( \dw t ) 	
\end{multline}
\begin{multline}
	\label{eq:bM}
	b_M(t) = 
		\big( b_{m}(t) + b_{m+1}(t) \big) \cos( \dw t ) 			
	\\
		 - \big( a_{m}(t) - a_{m+1}(t) \big) \sin( \dw t )  	
\end{multline}
and
\[
	\wM = (\omega_m + \omega_{m+1})/2,
	\quad
	\dw= (\omega_{m+1} - \omega_{m})/2.
\]
Here $\wM$ is the frequency midway between $ \om_m$ and $ \om_{m+1} $.
Equation \eqref{eq:gSM} expresses $ g_S(t) $ in terms of a single center frequency, $ \wM $,
instead of two distinct frequencies as in \eqref{eq:gST}.
Note that $ a_M(t) $ and $ b_M(t) $ are readily obtained from the time varying amplitudes $a_k(t)$ and $b_k(t)$.
The functions $ a_M(t) \cos(\om_M t) $ and $ b_M(t) \sin(\om_M t) $ are illustrated
in Fig.\ref{fig:Example_artifical_sfad_recover},
where $ a_M(t) $ and $ b_M(t) $ are shown as dashed lines.
Note that these amplitude functions are piecewise smooth (not piecewise constant),
due to the $ \cos(\dw t)$ and $ \sin(\dw t) $ terms.

To obtain an instantaneous phase function from \eqref{eq:gSM}
it is furthermore convenient to express $ g_S(t) $ in terms of  $ M(t) \exp(\myJ \, \om_M t) $.
To that end, we write $ g_S(t) $ as
\begin{multline}
	g_S(t) = 
	\half a_M(t) \, \myE^{\myJ \om_M t}
	+
	\half a_M(t) \, \myE^{-\myJ \om_M t}
	\\
	+
	\frac{1}{2\myJ} b_M(t) \, \myE^{\myJ \om_M t}
	-
	\frac{1}{2\myJ} b_M(t) \, \myE^{-\myJ \om_M t}
\end{multline}
or
\begin{multline}
	g_S(t) = 
	\left[	\half a_M(t) + \frac{1}{2\myJ} b_M(t) \right] \myE^{\myJ \om_M t}
	\\
	+
	\left[ \half a_M(t) - \frac{1}{2\myJ} b_M(t) \right] \myE^{-\myJ \om_M t}
\end{multline}
which we write as
\begin{equation}
	g_S(t) = \half \, g_+(t) \, \myE^{\myJ \om_M t} + \half \, \conj{[g_+(t)]} \, \myE^{-\myJ \om_M t}
\end{equation}
where $ g_+(t)$ is the `positive frequency' component,
defined as
\begin{equation}
	\label{eq:gpos}
	g_+(t) :=  a_M(t) + \frac{1}{\myJ} \, b_M(t).
\end{equation}
According to the model assumptions, $ g_+(t)$ is expected to be  piecewise smooth with the exception of sparse discontinuities.
We can use \eqref{eq:gpos} to define the instantaneous phase function
\begin{equation}
	\label{eq:thetaM}
	\theta_M(t) = -\tan\inv\biggl( \frac{ b_M(t) }{ a_M(t) } \biggr).
\end{equation}
The function $ \theta_M(t) $ represents the deviation of $ g_S(t) $ around its center frequency, $ \om_M $.
It is useful to use the four-quadrant arctangent function for \eqref{eq:thetaM}, i.e., `atan2'.
For the current example,
the instantaneous phase $ \theta_M(t) $
for the 9.5 Hz signal, $g_S(t)$, is illustrated in Fig.~\ref{fig:Example_artifical_sfad_recover}.
It clearly shows a discontinuity at $ t = 0.4 $ of about 180 degrees.

\begin{figure}
	\centering
		\includegraphics[scale = \figurescale] {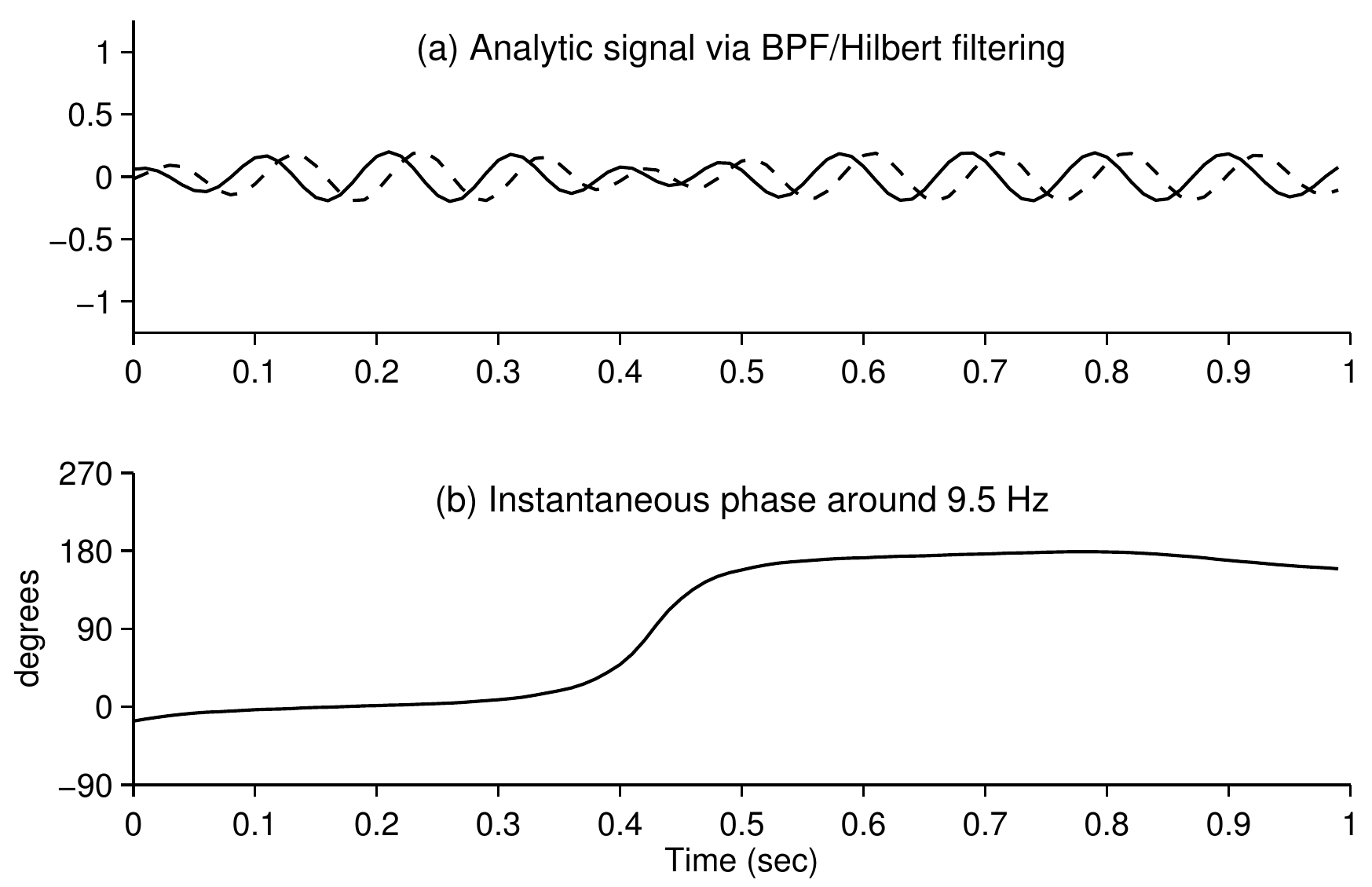} 
		\caption{
		Example 3: 
		The estimate of the 9.5 Hz component using LTI band-pass filtering.
		The instantaneous phase, computed using the Hilbert transform,
		exhibits a gradual phase shift.
		}
		\label{fig:Example_artifical_sfad_hilbert_phase}
\end{figure}

A finer frequency discretization would also be effective here to reduce the issue of
the narrow-band signal component (9.5 Hz) falling between discrete frequencies (9 and 10 Hz).
However, an aim of this example is to demonstrate how this case
is effectively handled when using sparse frequency analysis (SFA).

The estimate of the 9.5 Hz signal $ r(t) $ obtained by band-pass filtering
is illustrated in Fig.~\ref{fig:Example_artifical_sfad_hilbert_phase}a (solid line).
The utilized band-pass filter was designed to pass frequencies 8--12 Hz (i.e., alpha rhythms).
The Hilbert transform is also shown (dashed line).
The instantaneous phase of the analytic signal  (around 9.5 Hz)
is illustrated in Fig.~\ref{fig:Example_artifical_sfad_hilbert_phase}b.
It can be seen that the instantaneous phase
undergoes a 180 degree shift around $ t = 0.4 $ seconds, but the transition is not sharp.
Given a real-valued signal $ y(t) $, a complex-valued analytic signal $y_a(t)$ is formed 
by $ y_a(t) = y(t) + \myJ \, y_H(t) $, where
$ y_H(t) $ is the Hilbert transform of $ y(t) $.

In contrast with LTI filtering (band-pass filter and Hilbert transform),
the sparse frequency analysis (SFA) method yields an instantaneous phase function that 
accurately captures the step discontinuity at $ t = 0.4 $.
Hence, unlike LTI filtering, the sparse frequency analysis (SFA) approach
makes possible the high resolution estimation of phase discontinuities
of a narrow-band signal buried within a noisy multi-band signal.
This is true even when the center frequency of the narrow-band signal
falls between the discrete frequencies $ f_k $.

\section{Measuring Phase Synchrony}

The study of synchronization of various signals is of interest in biology, neuroscience, and in the study of the dynamics of physical systems \cite{Pikovsky_2002, Lachaux_1999}.
Phase synchrony is a widely utilized form of synchrony, which is thought to play a role in the integration of functional areas of the brain,
in associative memory, and motor planning, etc. \cite{Lachaux_1999,  Varela_1995, Tononi_1998}.
Phase synchrony is often quantified by the \emph{phase locking value} (PLV).
Various methods for quantifying, estimating, extending, and using phase synchrony have been developed \cite{Almeida_2011, Aviyente_2011, Sanqing_2010}.

The phase synchrony between two signals is meaningfully estimated when each signal is approximately narrow-band.
Therefore, methods to measure phase synchrony generally utilize band-pass filters designed to capture the frequency band of interest.
Generally, the Hilbert transform is then used to compute a complex analytic signal from which the time-varying phase is obtained.
In this example, we illustrate the use of SFA for estimating the instantaneous phase difference between two channels of a multichannel EEG.

\begin{figure}
	\centering
	\includegraphics{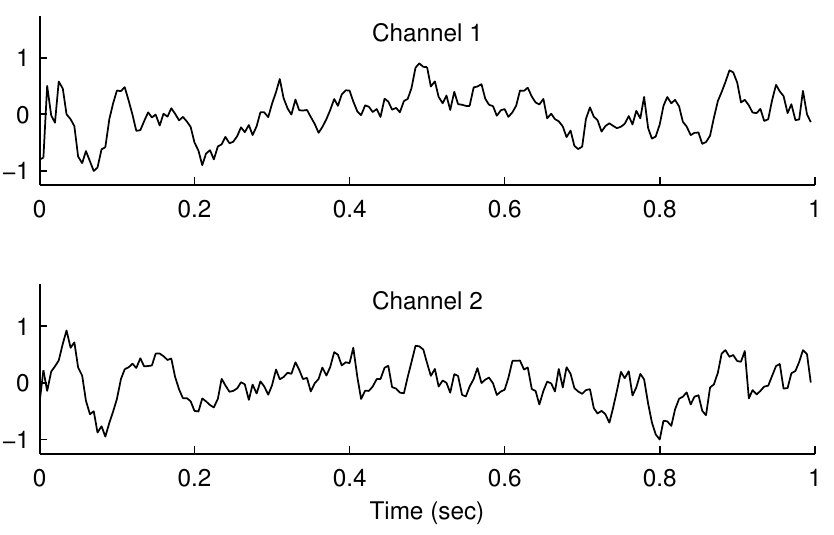} 
		\caption{
		Example 4: 
		Two channels of a multichannel EEG signal.
		}
		\label{fig:Example4_EEG}
\end{figure}

\subsection{Example 4: EEG instantaneous phase difference}

\begin{figure}[!t]
	\centering
		\includegraphics{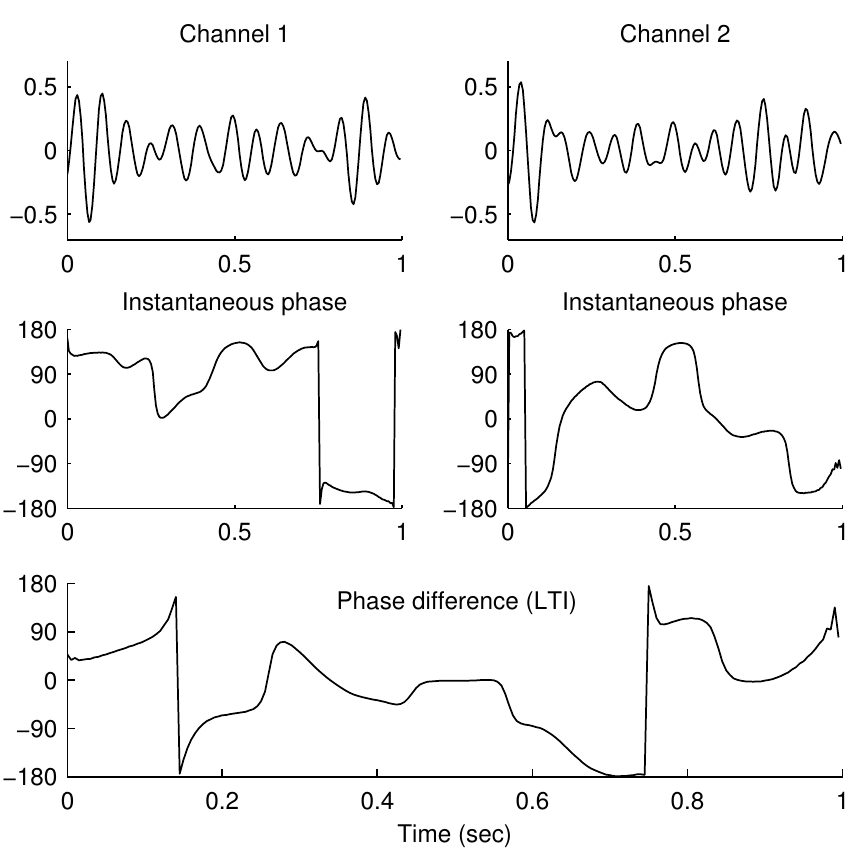} 
		\caption{
		Example 4:
		Band-pass (11-15 Hz) signals estimated using LTI band-pass filtering.
		Instantaneous phase functions obtained using Hilbert transform.
		Channel 1 (left) and channel 2 (right).
		Bottom: instantaneous phase difference.
		}
		\label{fig:Example4_LTI}
\end{figure}

Two channels of an EEG, with sampling rate $f_s = 200$ Hz,
are shown in Fig. \ref{fig:Example4_EEG}.
Band-pass signals in the 11--15 Hz band, obtained via band-pass filtering
are illustrated in Fig.~\ref{fig:Example4_LTI}.
The instantaneous phase functions (around 13 Hz)
are computed using the Hilbert transform, and the phase difference is shown.
The computation of the phase locking value (PLV) and other phase-synchronization indices
are based on the difference of the instantaneous phase functions.

The sparse frequency analysis (SFA) technique provides an alternative approach to obtain the instantaneous phase 
of each channel of the EEG and the phase difference function.
Phase synchronization indices can be subsequently calculated,
as they are when the instantaneous phase difference is computed via LTI filtering.
Here we use problem formulation \eqref{eqn:sfa_cost} with 
$ K = 100 $, with the frequencies $ f_k $, equally spaced from 0 to 99 Hz, i.e., $ f_k = k $ Hz, $0 \le k \le K-1$.
With this frequency grid, the 11--15 Hz band corresponds to five frequency components, i.e. $ k \in S = \{11, \dots, 15\}$.
The 11-15 Hz band can then be identified as $g_S(t)$ in \eqref{eq:gS}.

\begin{figure}	
	\centering
		\includegraphics{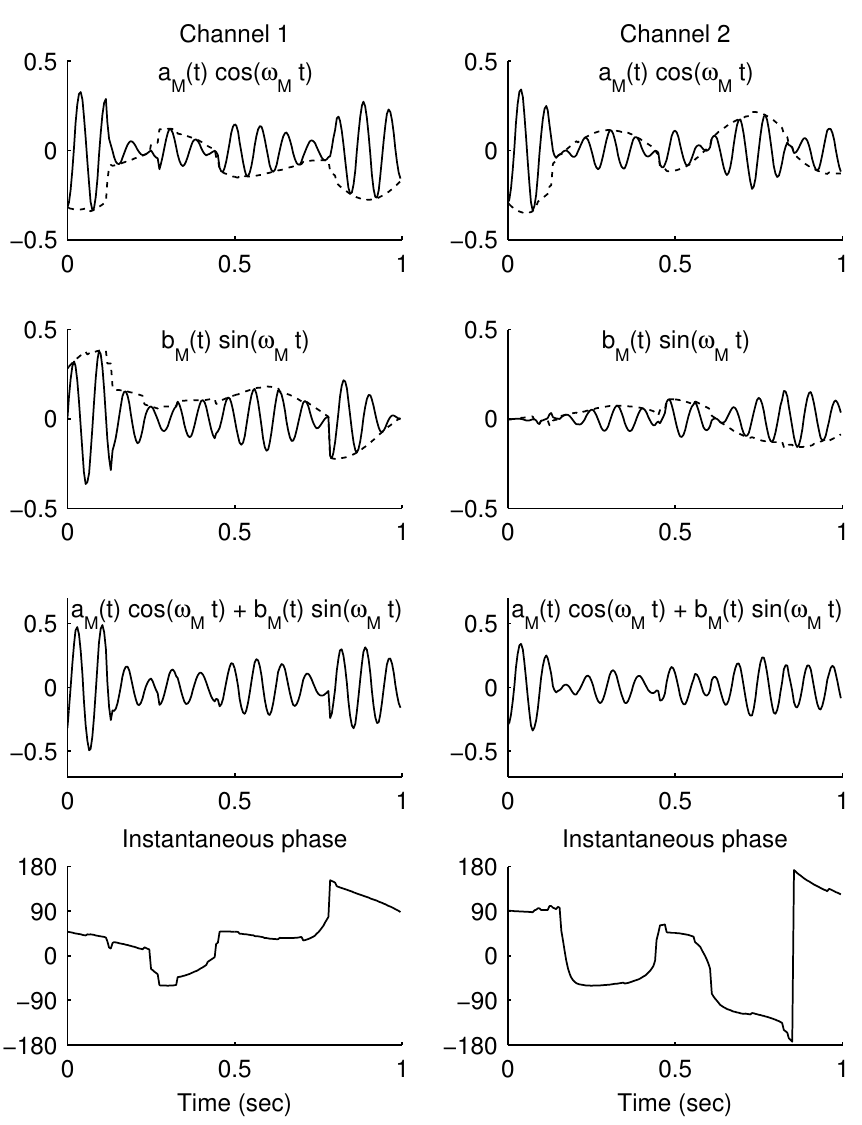} 
		\medskip
		\includegraphics{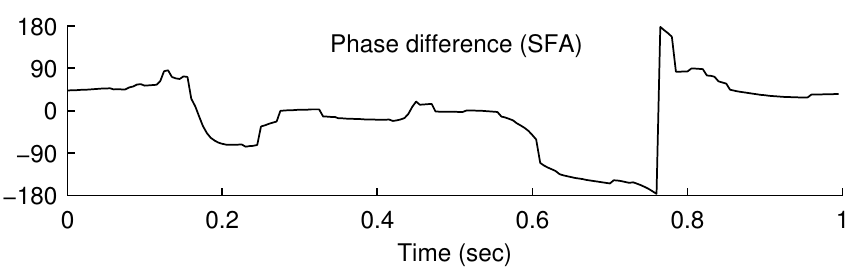} 
		\caption{
		Example 4:
		Band-pass (11-15 Hz) signals estimated using sparse frequency analysis (SFA).
		Channel 1 (left) and channel 2 (right).
		Bottom: instantaneous phase difference.
		}
		\label{fig:Example4_SFA}
\end{figure}

In order to compute the instantaneous phase of $g_S(t)$, we
express $g_S(t)$ in terms of sins/cosines with 
time-varying amplitudes and constant frequency as in \eqref{eq:gSM}.
For this purpose, we write $ g_S(t) $ as
\begin{equation}
	g_S(t) = g_{\{11, 15\}}(t) +  g_{\{12, 14\}}(t) +  g_{\{13\}}(t).
\end{equation}
Using \eqref{eq:aM} and \eqref{eq:bM}, we can write
\begin{equation}
	g_{\{13\}}(t) = 
	a^{(0)}_M(t) \cos(\om_M t) + b^{(0)}_M(t) \sin(\om_M t)
\end{equation}
\begin{equation}
	g_{\{12, 14\}}(t) = 
	a^{(1)}_M(t) \cos(\om_M t) + b^{(1)}_M(t) \sin(\om_M t)
\end{equation}
\begin{equation}
	g_{\{11, 15\}}(t) = 
	a^{(2)}_M(t) \cos(\om_M t) + b^{(2)}_M(t) \sin(\om_M t)
\end{equation}
where 
$ \om_M = 2\pi f_{13} = 26\pi $, with $f_{13}$ being the middle of the five frequencies in $S$.
The functions $ a^{(i)}_M(t)$, $b^{(i)}_M(t) $
are obtained by \eqref{eq:aM} and \eqref{eq:bM} from 
the amplitude functions $a_k(t)$, $b_k(t)$ produced by SFA.
Therefore, we can write the 11-15 Hz band
signal, $ g_S(t) $, as \eqref{eq:gSM}
where 
\begin{equation}
	a_M(t) = a^{(0)}_M(t)
	+
	a^{(1)}_M(t)
	+
	a^{(2)}_M(t)
\end{equation}
\begin{equation}
	b_M(t) = b^{(0)}_M(t)
	+
	b^{(1)}_M(t)
	+
	b^{(2)}_M(t).
\end{equation}
Further, the instantaneous phase of $ g_S(t) $ can be obtained using \eqref{eq:thetaM}.
The 11-15 Hz band of each of the two EEG channels
so obtained via SFA, 
and their instantaneous phase functions, are illustrated in Fig.~\ref{fig:Example4_SFA}.

Comparing the phase difference functions obtained by SFA and BPF/Hilbert (LTI) filtering,
it can be noted that they are quite similar
but the phase difference estimated by SFA is somewhat more stable.
Between 0.3 and 0.4 seconds,
the SFA phase-difference varies less than the LTI phase-difference.
Moreover, 
in the interval between 0.15 and 0.2 seconds,
the LTI phase difference is increasing,
while for SFA it is decreasing.

The true underlying subband signals are unknown;
yet, if they do possess abrupt amplitude/phase transitions,
the SFA technique may represent them more accurately.
In turn, SFA, may provide more precise timing of phase locking/unlocking.

\section{Conclusion}

This paper describes a sparse frequency analysis (SFA) method by which an $N$-point discrete-time signal $x(n)$
can be expressed as the sum of sinusoids wherein the amplitude
and phase of each sinusoid is a time-varying function.
In particular, the amplitude and phase of each sinusoid is modeled as
being approximately piecewise constant (i.e., the temporal derivatives of the instantaneous amplitude and phase functions are modeled as sparse).
The signal $x(n)$ is furthermore assumed to have a sparse frequency spectrum
so as to make the representation well posed.
The SFA method can be interpreted as a generalization of the discrete Fourier transform (DFT),
since, highly regularizing the temporal variation of the amplitudes of the sinusoidal components
leads to a solution wherein the amplitudes are non-time-varying.

The SFA method, as described here, is defined by a convex optimization problem,
wherein the total variation (TV) of the sine/cosine amplitude functions,
and the frequency spectrum are regularized,
subject to either a perfect or approximate reconstruction constraint.
An iterative algorithm is derived using variable splitting, ADMM, and MM
techniques from convex optimization.
Due to the convexity of the formulated optimization problem and the properties of ADMM, the algorithm converges reliably to the unique optimal solution.

Examples showed that the SFA technique can be used to perform mildly non-linear band-pass filtering
so as to extract a narrow-band signal from a wide-band signal, even
when the narrow-band signal exhibits amplitude/phase jumps.
This is in contrast to conventional linear time-invariant (LTI) filtering
which spreads amplitude/phase discontinuities across time and frequency.
The SFA method is illustrated using both synthetic signals and human EEG data.

Several extensions of the presented SFA method are envisioned.
For example, depending on the data, a non-convex formulation that more strongly promotes sparsity may be suitable.
Methods such as re-weighted L1 or L2 norm minimization \cite{Candes_2008_JFAP,Wipf_2010_TSP, Gorodnitsky_1997_TSP} or greedy algorithms \cite{Mallat_1993} can be used
to address the non-convex  form of SFA.
In addition, 
instead of modeling the frequency components as being approximately piecewise constant,
it will be of interest to model them as being piecewise smooth.
In this case, the time-varying amplitude functions may be regularized
using generalized or higher-order total variation \cite{Bredies_2010_SIAM,Karahanoglu_2011_TSP,Hu_2012_TIP,Rodriguez_2009_TIP}, or a type of wavelet transform \cite{BGG97}.
Incorporating higher-order TV into the proposed SFA framework 
is an interesting avenue for further study.

\bibliographystyle{plain}

\end{document}